\pdfoutput=1
\documentclass[letterpaper, 10 pt, conference]{ieeeconf}
\IEEEoverridecommandlockouts
\usepackage[utf8]{inputenc}
\usepackage{times}
\usepackage{amsmath}
\usepackage{amssymb}
\usepackage{graphicx}
\usepackage{booktabs}
\usepackage{multirow}
\usepackage{array}
\usepackage{xcolor}
\usepackage{url}
\usepackage{tikz}
\usetikzlibrary{positioning,arrows.meta,fit,calc}
\makeatletter
\let\NAT@parse\undefined
\makeatother
\usepackage[hidelinks]{hyperref}
\usepackage{orcidlink}
\makeatletter
\AtBeginDocument{%
  \def\@citex[#1]#2{%
    \leavevmode
    \let\@citea\@empty
    \@cite{\@for\@citeb:=#2\do
      {\@citea\def\@citea{], [}%
       \edef\@citeb{\expandafter\@firstofone\@citeb\@empty}%
       \if@filesw\immediate\write\@auxout{\string\citation{\@citeb}}\fi
       \@ifundefined{b@\@citeb}{\mbox{\reset@font\bfseries ?}%
         \G@refundefinedtrue
         \@latex@warning{Citation `\@citeb' on page \thepage\space undefined}}%
         {\hyper@@link[cite]{}{cite.\@citeb}{\csname b@\@citeb\endcsname}}}}{#1}}}
\makeatother

\DeclareMathOperator{\softmax}{softmax}

\newcommand{\ours}{HAVM}
\newcommand{\pp}{\,pp}
\newcolumntype{L}[1]{>{\raggedright\arraybackslash}p{#1}}
\newcolumntype{C}[1]{>{\centering\arraybackslash}p{#1}}

\title{\LARGE \bf
Hetero-Associative Visuomotor Memory: Soft Recall as a Base\\for Learned Residuals in Few-Demonstration Imitation
}

\author{Runcong Wang\,\orcidlink{0009-0006-1019-9481}%
\thanks{Runcong Wang received the Master's degree from the Technical University of Munich.}}

\begin{document}
\maketitle
\thispagestyle{empty}
\pagestyle{empty}

\begin{abstract}
Modern visuomotor policies (ACT, diffusion policies, and VLAs) use tens of millions to billions of parameters and hours
of training, yet on small datasets large models like diffusion policies act largely like an implicit action memory.
Retrieve-and-correct policies make this memory explicit and refine a retrieved chunk with a learned corrector, usually
starting from the nearest neighbor. We show that on sparse, varied human demonstrations, a smooth softmax average over
the whole memory is a better base for correction than the nearest chunk, even when it performs worse alone. On ALOHA
cube transfer (40 human demonstrations), correcting soft recall beats top-1 by 22.4 points on fresh encoders and resets
(7.4 against a tuned top-1 corrector), mainly as a steadier additive anchor. On robomimic Square, it leads by 7.9 points
on two better operators and 6.7 on an equal mixed draw; accuracy depends on data quality, and the gain fades on dense,
scripted, or generated data. Our hetero-associative visuomotor memory (\ours{}, 14\,M parameters) trains offline in five
minutes via leave-one-demonstration-out recall and runs in 1.1\,ms, reaching 76.4\,\% in 24\,s across six ALOHA encoders
and leading all 40-demo baselines in 8\,s (74.6\,\%). Averaging three seeded heads on the shared encoder and memory
removes single-head blow-ups: ALOHA reaches 79.5\,\% (24\,s) and 69.1\,\% strict (Diffusion Policy 78.8\,\% and
70.2\,\%), Threading gains 8.8 points (7.0 on its strongest variant, to 68.3\,\%), and MimicGen Square gains 7.7, at
1.5\,ms (2.8\,ms without CUDA graphs; ACT 3.5\,ms) and under 200\,MB GPU memory. One 40-task LIBERO memory reaches
92.2\,\% (8\,ms) with smaller rewording drops than SmolVLA, and a 6\,ms re-plan sustains 30\,Hz control on a real SO-101
arm. \ours{} targets fast adaptation on consistent workspaces and single-strategy tasks.
\end{abstract}

\section{Introduction}
\label{sec:intro}
\begin{figure*}[t]
  \centering
  \includegraphics[width=\textwidth]{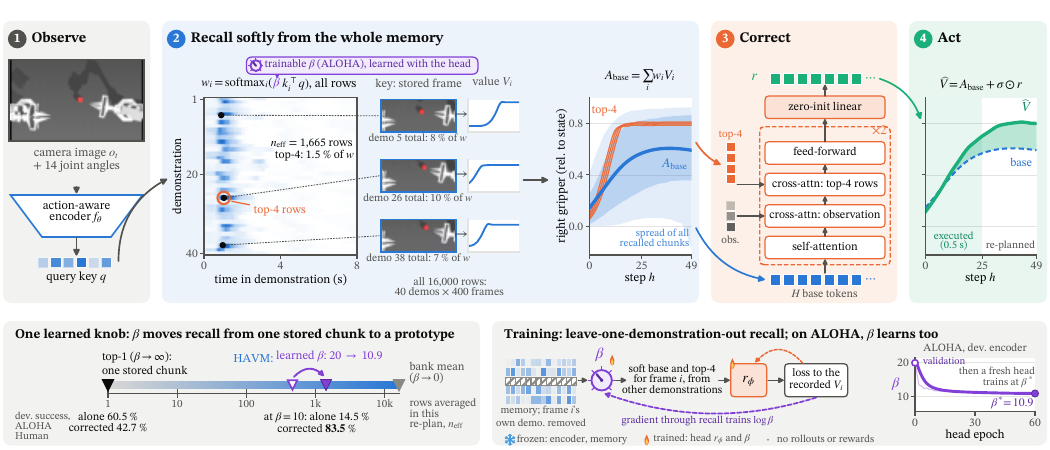}
  \caption{\ours{} at one re-plan of a test rollout (ALOHA cube transfer, 40 human demonstrations, development encoder at
  fixed $\beta=10$, $t=1$\,s). Panels 1, 2, and 4 and the $\beta$ curve show stored data; panel 3 and the training
  diagram are schematic. (1)~The encoder maps the image and joint state to a key. (2)~A softmax over all 16\,000 rows,
  controlled by the inverse temperature $\beta$ (violet), spreads weight across the same phase of many demonstrations
  ($n_{\mathrm{eff}}=1\,665$ rows in this re-plan; median over re-plans is 2\,454, Table~\ref{tab:beta}). (3)~A
  2.8\,M-parameter attention head reads this soft base, the top-4 rows, and the observation to predict a correction.
  (4)~The corrected chunk. Bottom: $\beta$ moves recall from one stored chunk to the bank mean~\cite{wangHASM}. The head
  trains on recall with the frame's whole demonstration removed. In the offline $\beta$ rule, the same loss jointly
  updates $\log\beta$ from the validation choice (20) to $\beta^*=10.9$, close to the closed-loop $\beta=10$ illustrated
  above, after which a fresh head is trained at $\beta^*$.}
  \label{fig:overview}
\end{figure*}

Deploying modern robotic systems is often resource-intensive, requiring high-performance computing hardware, extensive
demonstration data, long training times, and substantial power budgets. Widely used policies, including action-chunking
transformers~\cite{zhaoACT2023}, diffusion policies~\cite{chiVisuomotorPolicyLearning}, and vision-language-action
models (VLAs)~\cite{blackPi02024,kimOpenVLA2024,shukorSmolVLA2025,zhengXVLA2025}, use tens of millions to billions of
parameters and hours of training or fine-tuning. When the workspace is consistent, the tasks are straightforward, and
demonstrations are sparse, large models waste compute and add unnecessary latency. He et
al.~\cite{heDemystifyingDiffusionPolicies2025} found that diffusion policies trained on small datasets largely
reproduce training sequences and act as an implicit lookup table. An explicit action lookup table (ALT) matches them
far more cheaply, as our own Diffusion Policy experiments confirm (Appendix~\ref{app:baselines}).

We therefore build the lookup table explicitly as an associative memory that can be inspected and queried in
milliseconds. In our earlier work~\cite{wangHASM}, a hetero-associative sequential memory recalled the next motor
pattern via the modern Hopfield update $Z=\softmax(\beta QK^{\top})V$~\cite{ramsauerHopfieldNetworksAll2021}, where
$\beta$ interpolates between one stored pattern and a prototype over several. Because lookup can only replay recorded
motions, a policy also needs to adapt when no demonstration matches the scene. Retrieve-and-correct
methods~\cite{sridharMCNN2024,sridharREGENT2025,parkReCAP2026,pfeiferDARP2026,adangBPC2026} add a learned network to
refine retrieved actions, usually passing it the nearest neighbors that retrieval alone would execute.

Our study shows that nearest neighbors are not the best base for a corrector when demonstrations are sparse and varied.
Noise in a single demonstration can mislead the corrector, and the matched demonstration can switch abruptly from one
chunk to the next.

Our hetero-associative visuomotor memory, \ours{} (Fig.~\ref{fig:overview}), applies this finding. An action-aware
ResNet encoder maps frames with similar futures close together, and the memory stores one key and one future action
chunk per frame. Soft recall forms a draft chunk, and a 2.8\,M-parameter attention head reads that draft as an anchor,
the top-4 nearest neighbors as fine-grained references, and the observation to predict an additive correction. Training
uses \emph{leave-one-demonstration-out} recall so the head learns from recall errors on unseen demonstrations, matching
larger policies with less training time, fewer parameters, and lower latency. In summary:
\begin{itemize}
  \item We test what base a corrector should receive with all else fixed. On ALOHA Human, the soft base beats the
  nearest chunk by 22.4 points on fresh resets across three encoder and three head seeds. This advantage persists
  against a closed-loop-tuned top-1 baseline and across different corrector architectures, and comes mainly from using
  the soft base as the additive anchor. On robomimic Square, it leads by 7.9 points with the two better operators and by
  6.7 on a label-free draw of equal size. Confirmed on additional data and a controlled toy, the gain appears on sparse,
  varied demonstrations and fades on dense or scripted data.
  \item We present \ours{}, a 14\,M-parameter policy trained offline in about five minutes (with an offline learned
  $\beta$) with a 1.1\,ms call. Across six fresh encoders, it reaches 76.4\,\% at 24\,s, comparable to a
  264\,M Diffusion Policy trained for 6.7\,h and ahead of ACT and fine-tuned VLAs, and reaches 74.6\,\% within 8\,s,
  leading every 40-demonstration baseline ($+6.5$ over Diffusion Policy, $+15.1$ over ACT). Averaging three
  independently seeded heads over the shared encoder and memory removes single-head blow-ups: ALOHA reaches 79.5\,\% at
  24\,s and 69.1\,\% strict (Diffusion Policy 78.8 and 70.2\,\%), Threading gains 8.8 points (7.0 on its strongest
  variant, to 68.3\,\%), and MimicGen Square gains 7.7, with a 1.5\,ms call (2.8\,ms without CUDA graphs; ACT 3.5\,ms)
  and 186--193\,MB of GPU memory (ACT 266\,MB). On LIBERO, one memory reaches 92.2\,\% and loses far less to rewording
  than SmolVLA; on a real SO-101 arm, its 6\,ms re-plan sustains 30\,Hz control without stalling.
  \item We examine the limitations of retrieval-based policies, including multi-modal trajectory averaging and failures
  at first contact in areas with few demonstrations.
\end{itemize}

\section{Related Work}
\label{sec:related}
Modern visuomotor imitation learning usually maps observations to actions with large parametric models: action chunking
shortens the decision horizon to curb compounding errors~\cite{zhaoACT2023,rossEfficientReductionsImitation,rossDAgger2011},
diffusion and flow-matching policies capture multi-modal
actions~\cite{chiVisuomotorPolicyLearning,lipmanFlowMatching2023,blackPi02024}, and VLAs build on pretrained
vision-language backbones~\cite{blackPi02024,kimOpenVLA2024,kimOpenVLAOFT2025,shukorSmolVLA2025,zhengXVLA2025}. In
low-data experiments, He et al.~\cite{heDemystifyingDiffusionPolicies2025} found that Diffusion Policy often generated
action chunks closely matching training demonstrations, suggesting a form of implicit action memory.

Retrieval policies make this memory explicit and execute the readout directly.
ALT~\cite{heDemystifyingDiffusionPolicies2025} executes the top-1 chunk under a contrastive key,
VINN~\cite{pariVINN2022} averages $k$ neighbors, locally weighted learning~\cite{atkesonLocallyWeightedLearning1997}
averages nearby experiences, and Grachev et al.~\cite{grachevKNNVLA2026} average 13 neighbors on LIBERO. Associative
memories span this range with one parameter: modern Hopfield
networks~\cite{krotovDenseAssociativeMemory2016,ramsauerHopfieldNetworksAll2021} cast retrieval as attention whose
inverse temperature $\beta$ moves fixed points between single patterns, metastable averages, and the global mean.
Hetero-associative variants return a distinct target pattern, such as the next step in a
sequence~\cite{koskoBidirectionalAssociativeMemories1988,chaudhryLongSequenceHopfield2023}. Our earlier
work~\cite{wangHASM} used this sequential form for mobile manipulation; \ours{} applies the same update over a learned
key across the whole memory.

Executing a retrieved readout directly, however, can only replay or blend recorded demonstrations. Refining that
prototype with a learned residual appears across several fields. In nonparametric
regression, a softmax over all stored rows is the Nadaraya--Watson
estimator~\cite{nadarayaRegression1964,watsonSmoothRegression1964} in a learned metric, leave-one-out is standard for
fitting kernel smoothers~\cite{stoneCrossValidation1974,hardleMarronBandwidth1985}, and smoothing a smoother's residuals
is Tukey's twicing~\cite{tukeyEDA1977}, formalized by L2-boosting~\cite{buhlmannL2Boost2003}. Classical learning from
demonstration likewise encodes a mean trajectory and its variability before adapting them, as in DMPs~\cite{ijspeertDMP2013}, ProMPs~\cite{paraschosProMP2013}, and
task-parameterized GMMs~\cite{calinonTPGMM2016}, while BeT~\cite{shafiullahBeT2022} and VQ-BeT~\cite{leeVQBeT2024} predict a
cluster or codebook center plus a learned offset. None of these studies test
whether a smooth visuomotor readout beats the nearest neighbor as a base even when it performs worse on its own.

\begin{table}[t]
\caption{Retrieve-and-correct methods differ in what the learner receives as its base.}
\label{tab:family}
\centering\footnotesize
\setlength{\tabcolsep}{3pt}
\begin{tabular}{@{}L{1.65cm}L{2.65cm}L{3.5cm}@{}}
\toprule
Method & Base & Correction \\
\midrule
MCNN~\cite{sridharMCNN2024} & nearest Neural-Gas memory & bounded net, blended by $e^{-\lambda d}$ \\
REGENT~\cite{sridharREGENT2025} & 1-NN action, 19 neighbors read & transformer, blended with 1-NN \\
ReCAP~\cite{parkReCAP2026} & nearest chunk & residual from a frozen policy \\
DARP~\cite{pfeiferDARP2026} & $k$ neighbors & mean of neighbor proposals \\
BPC~\cite{adangBPC2026} & signed sum of windows & closed-form ridge residual \\
RL4IL~\cite{ismkhanRL4IL2026} & fusion of 32 labels & MLP on the fused label \\
\midrule
\ours{} & softmax over the whole bank & attention head, leave-one-demo-out \\
\bottomrule
\end{tabular}
\end{table}

Beyond these prototype models, modern robotic policies also use retrieval and residual correction in several ways.
Retrieved data can train a
policy~\cite{duBehaviorRetrieval2023,linFlowRetrieval2024,memmelSTRAP2025,kumarCOLLAGE2025,xieIWR2025} or condition a
large policy at test time~\cite{sridharRICL2025,zhangRetrievalVLA2026,zhuRAEA2024,papagiannisRX2025,vosyliusInstantPolicy2025},
while residual learning over a base controller is standard in reinforcement
learning~\cite{silverResidualPolicy2018,johanninkResidualRL2019,ankileResiP2024,yuanPolicyDecorator2024}. The methods
closest to ours combine both by refining retrieved actions with a learned network (Table~\ref{tab:family}). Their bases
range from the nearest neighbor to an average over $k$ neighbors (DARP) or a fusion of 32 ranked labels, which RL4IL
finds to match or outperform the single best candidate~\cite{ismkhanRL4IL2026}. In contrast, we vary only the base, from
the nearest chunk to a softmax over the whole memory, while holding the key, candidates, and training fixed.

\section{Hetero-Associative Visuomotor Memory}
\label{sec:method}
\subsection{Problem and memory}
\label{sec:problem}
We are given $D$ demonstrations $\mathcal D_e=\{(o_{e,t},a_{e,t})\}_{t=0}^{T_e-1}$ ($e=1,\dots,D$). Demonstration $e$ has
$T_e$ frames indexed by $t$, with observation $o_{e,t}$ and controller command $a_{e,t}\in\mathbb R^{J}$ ($J$ action
dimensions); we drop $e$ when clear. An observation $o_t$ holds the camera images, measured $S$-dimensional robot state
$s_t$, and, for multi-task data, an instruction $\ell$. The policy predicts an $H$-step chunk, executes the first $E<H$
steps, and re-plans~\cite{zhaoACT2023}.

The memory stores one row per demonstration frame, indexed by $i=(e,t)$, totaling $N=\sum_eT_e$ rows. Row $i$'s value is
the chunk following that frame,
\begin{equation}
 A_i[h]=a_{e,\min(t+h,\,T_e-1)},\qquad h=0,\dots,H-1,
 \label{eq:chunk}
\end{equation}
where $h$ indexes the chunk step and the end is padded with the last command. This extends the back-shifted value matrix
of our sequential memory~\cite{wangHASM} from one next pattern to the next $H$ commands. For ALOHA's joint-position
commands, we store relative targets $V_i[h]=A_i[h]-s_i$, with $s_i$ the state at frame $i$; for the Panda benchmarks, we
store operational-space commands (translation, rotation, and gripper) as recorded, $V_i=A_i$. Row $i$'s key is
$k_i=f_\theta(o_i)$, produced by a learned encoder $f_\theta$ (Section~\ref{sec:key}). Recall applies the
hetero-associative modern Hopfield update,
\begin{equation}
 w=\softmax_i(\beta\gamma_i),\qquad A_{\mathrm{base}}=\sum_{i=1}^{N}w_iV_i,\qquad \gamma_i=k_i^{\top}q,
 \label{eq:readout}
\end{equation}
where $q=f_\theta(o)$ is the query for observation $o$, $\gamma_i$ is row $i$'s similarity, and $w_i$ is its weight over
the \emph{whole} bank. The recalled chunk $A_{\mathrm{base}}$ is the \emph{base} corrected by the head. The inverse
temperature $\beta>0$ sets the regime: large $\beta$ selects one stored chunk, whereas small $\beta$ yields a prototype
over many rows. We measure this regime by the effective number of averaged rows, $n_{\mathrm{eff}}=\exp(-\sum_iw_i\log
w_i)$, from 1 (top-1) to $N$ (bank mean).

\subsection{Action-aware keys}
\label{sec:key}
The encoder follows ALT~\cite{heDemystifyingDiffusionPolicies2025}. A shared ImageNet-initialized
ResNet-18~\cite{heResNet2016} pools each camera image to 512 features, projected by a shared linear map to 128. The
standardized state passes through an $S{\to}64{\to}32$ MLP; for multi-task data, a frozen text encoder
(Appendix~\ref{app:config}) is projected to 32 similarly. A fusion MLP (256, then 128) outputs the unit-norm key; all
MLPs use ReLU. Training uses two augmented views per frame (random resized crops, color jitter, and state noise) with
two losses. SimCLR's NT-Xent~\cite{chenSimCLR2020} aligns both views, while an \emph{action-aware} term aligns frames
with similar future chunks. For frame $i$ in a batch, let $\widetilde V_i$ be its value standardized per action dimension
and $\mathcal E_i$ the batch frames $j$ from other demonstrations. The loss is the cross-entropy between a target
distribution $p_{ij}$ from future-action similarity and $\hat p_{ij}$ from the keys:
\begin{equation}
\begin{aligned}
 p_{ij}&=\softmax_{j\in\mathcal E_i}\!\big(-\|\widetilde V_i-\widetilde V_j\|_1/(HJ\,\sigma_a)\big),\\
 \hat p_{ij}&=\softmax_{j\in\mathcal E_i}\!\big(z_i^{\top}z_j'/\tau_a\big),
\end{aligned}
 \label{eq:actionaware}
\end{equation}
where $z_i$ and $z_j'$ are the keys of view 1 of frame $i$ and view 2 of frame $j$, the L1 distance is averaged over $HJ$
chunk entries, and $\sigma_a=\tau_a=0.1$. The encoder is selected on validation demonstrations and frozen, then every
training frame is encoded once into the bank. We define the \emph{validation $\beta$} as the inverse temperature chosen
over a grid ($10$--$1\,000$) to minimize the chunk error of uncorrected soft recall on held-out validation
demonstrations. In early tests without the head, this term made soft recall viable on ALOHA Human (70\,\% within 8\,s
against 0\,\% with SimCLR alone, at the validation $\beta$; Appendix~\ref{app:ablations}).

\emph{Language.} When instructions vary, the text feature is fused into the key, the score gains a binding term
$\gamma_i=k_i^{\top}q+\lambda_\ell(u(\ell_i)^{\top}u(\ell)-1)$, and the head receives the text feature as an observation
token, where $u(\cdot)$ is the unit-norm text feature, $\ell_i$ row $i$'s instruction, and $\lambda_\ell$ the binding
weight. On LIBERO-Goal, this raised soft recall from 8 to 94\,\%.

\subsection{Residual attention head}
\label{sec:head}
The head $r_\phi$ corrects the base in standardized action units,
\begin{equation}
 \widehat V=A_{\mathrm{base}}+\sigma\odot r_\phi\big(\tilde b,\,\mathcal C,\,\mathcal O\big),
 \qquad \tilde b=(A_{\mathrm{base}}-\mu)/\sigma,
 \label{eq:residual}
\end{equation}
where $\widehat V$ is the predicted chunk, $\mu,\sigma\in\mathbb R^{J}$ are the per-dimension mean and standard deviation
of stored values (applied at every chunk step), $\odot$ is the element-wise product, and $\mathcal C$, $\mathcal O$ are
candidate and observation tokens. All tokens have width 256, and each $\mathrm{Lin}$ is a distinct learned linear map.
Each step $h$ of $\tilde b$ forms a query token $Q_h=\mathrm{Lin}(\tilde b_h)+\pi^{\mathrm{step}}_h$ with learned step
embedding $\pi^{\mathrm{step}}_h$. The candidates are the $K_c=4$ highest-scoring rows ($m=1,\dots,K_c$). With $\tilde
c_m$ candidate $m$'s standardized chunk and $\tilde w_m$ its softmax weight renormalized over the four, each candidate
step becomes $M_{m,h}=\mathrm{Lin}(\tilde c_{m,h})+\pi^{\mathrm{step}}_h+\pi^{\mathrm{rank}}_m+\mathrm{Lin}(\tilde
w_m)$, with learned rank embedding $\pi^{\mathrm{rank}}_m$, supplying detail removed by averaging. Observation tokens
include each camera feature, state feature, fused key, and optional text feature. Two blocks update the query tokens via
self-attention, cross-attention to observation and $K_cH$ candidate tokens, and a feed-forward layer (pre-norm, 8 heads,
dropout 0.1). Attention here is the modern Hopfield update with learned projections~\cite{ramsauerHopfieldNetworksAll2021}.
A \emph{zero-initialized} final linear layer outputs $r\in\mathbb R^{H\times J}$, so the untrained policy equals recall.
At each re-plan, the policy encodes the observation, computes $\gamma$ against all $N$ keys, forms the base and
top-$K_c$ candidates, and runs the head in one deterministic forward pass; for ALOHA, joint-position commands are
reconstructed from the current state $s$ as $\widehat A[h]=s+\widehat V[h]$. Together, encoder and head total 14\,M
parameters (2.8\,M in the head), and the 40-demonstration ALOHA bank holds 16\,000 rows (13\,M numbers). Each re-plan
costs one encoder pass, one $N\times128$ product, and one head pass independent of $N$.

\emph{Head ensemble.} Optionally, $M=3$ heads trained with different seeds on the same frozen encoder and memory are
averaged, $\widehat V=\frac1M\sum_{m=1}^{M}\big(A_{\mathrm{base}}^{(m)}+\sigma\odot r_{\phi_m}\big)$, where
$A_{\mathrm{base}}^{(m)}$ is recall at head $m$'s $\beta$; with a shared $\beta$ this simplifies to
$A_{\mathrm{base}}+\frac{\sigma}{M}\sum_m r_{\phi_m}$ (Section~\ref{sec:exp-when}). Because a batch-1 call is
launch-bound on the GPU, the ensemble shares the encoder pass and similarity scores and runs all heads over stacked
weights with one batched operation per layer. At 19.8\,M parameters, its ALOHA call takes 2.78\,ms p95 (one head 2.53, ACT 3.53\,ms), or 1.54\,ms with CUDA graphs (1.07\,ms), using 186--193\,MB of GPU memory (one head 132--139,
ACT 266\,MB; Appendix~\ref{app:baselines}). Heads with their own vision branch (Section~\ref{sec:exp-main}) are averaged
one by one (5.8 against 2.1\,ms).

\subsection{Leave-one-demonstration-out training}
\label{sec:training}
The head regresses the recorded chunk. If a training frame could recall its own demonstration, overlapping neighboring
chunks would provide near-copies of the target, teaching the head to trust nearly exact recall that never occurs at test
time. We therefore recall each training frame $i$ with its \emph{whole} demonstration $e(i)$ excluded,
\begin{equation}
 w^{(i)}_j=\softmax_{j:\,e(j)\neq e(i)}\big(\beta\gamma_j\big),
 \label{eq:mask}
\end{equation}
where $e(j)$ is row $j$'s demonstration and $\gamma_j$ its score for frame $i$'s key. The top-$K_c$ candidates come from
the same remaining rows, so both base and candidates originate from other demonstrations. With $A^{(i)}_{\mathrm{base}}$
recalled under $w^{(i)}$ and the encoder and bank frozen, we minimize
\begin{equation}
 \mathcal L(\phi)=\frac{1}{HJ}\Big\|\frac{A^{(i)}_{\mathrm{base}}-V_i}{\sigma}+r_\phi\Big\|_F^2 ,
 \label{eq:loss}
\end{equation}
averaged over training frames, where $V_i$ is frame $i$'s target chunk, $r_\phi$ reads frame $i$'s base, candidates, and
observation, and $\|\cdot\|_F$ is the Frobenius norm. Each stored frame thus forms a training pair matching recall's
error on an unseen demonstration. Without this exclusion, corrected soft recall on ALOHA Human falls from 83.5 to
60.5\,\% (Section~\ref{sec:exp-mechanism}). By contrast, BPC~\cite{adangBPC2026} excludes only same-demonstration
candidates overlapping in time. The head trains for 60 epochs (Appendix~\ref{app:config}); among EMA checkpoints saved
every ten epochs, we select the lowest chunk error on held-out validation demonstrations, which never enter the bank.

\subsection{Selecting $\beta$}
\label{sec:execution}
The sole retrieval hyperparameter is the inverse temperature $\beta$, which sets the sharpness of the base readout
$A_{\mathrm{base}}$. We consider three ways to choose $\beta$: offline validation on uncorrected recall, offline
$\beta$ learning through the corrector loss, and a closed-loop $\beta$ sweep.

\emph{Offline validation.} As defined in Section~\ref{sec:key}, the validation $\beta$ minimizes the held-out chunk
error of uncorrected soft recall over a grid ($10$--$1\,000$). Although it requires no rollouts or head training,
purely offline selection on uncorrected recall is inaccurate for a corrected policy: it systematically favors readouts
that are too sharp for the head, and offline chunk error misranks closed-loop settings (Section~\ref{sec:exp-when}).

\emph{Learned $\beta$.} To account for the corrector offline, we can instead optimize $\beta$ through the
leave-one-demonstration-out loss~\eqref{eq:loss} (Fig.~\ref{fig:overview}, bottom). Starting from the validation
$\beta$, $\log\beta$ trains jointly with a head for 60 epochs, and a fresh head is then trained at the epoch-60 EMA
value $\beta^*$. In the joint run, the head peaks near epoch 10 while $\beta$ is still moving, whereas by epoch 60
$\beta^*$ has settled and shifts by at most 1.5 across starts 20 and 40. On ALOHA Human this closes most of the gap to
closed-loop tuning, though offline learning does not transfer to every benchmark (Section~\ref{sec:exp-when}).

\emph{Closed-loop $\beta$ sweep.} Because purely offline selection is not reliable across all benchmarks, we can sweep
candidate $\beta$ values, typically softer than the validation choice, in closed loop. Since the encoder and bank stay
frozen, each candidate only requires retraining the lightweight head (Table~\ref{tab:config}; e.g., halving the
validation $\beta$ on ALOHA or using $\beta=1$ on MimicGen). Where the validation $\beta$ already suffices or
closed-loop sweeps are impractical (e.g., LIBERO, robomimic, and the real SO-101 arm), we retain the validation
$\beta$.

\section{Experiments}
\label{sec:exp}
We evaluate \ours{} around five questions. \textbf{Q1:} what readout should the memory supply to the corrector? \textbf{Q2:} how does \ours{}
compare against parametric and retrieval-based policies trained on the same demonstrations, in success and cost? \textbf{Q3:} does it carry over to multi-task suites, reworded instructions, and a real robot?
\textbf{Q4:} when do its design choices matter? \textbf{Q5:} what are \ours{}'s limitations, and how do they trade off against its strengths?

\subsection{Setup}
\label{sec:setup}
We evaluate \ours{} on five simulated and real-world manipulation benchmarks. \emph{ALOHA cube transfer}~\cite{zhaoACT2023} uses two arms, 14 joint commands, and one top
camera: the right arm lifts a cube and hands it to the left arm. It includes 50 human (\emph{Human}) and 50
scripted (\emph{Scripted}) demonstrations, of which we train on 40. \emph{robomimic} Can and
Square~\cite{mandlekarRobomimic2021} provide 200 human demonstrations each, \emph{MimicGen}~\cite{mandlekarMimicGen2023}
Square, Stack, Threading, and Coffee provide generated demonstrations, \emph{LIBERO}~\cite{liuLIBERO2023} contains four
suites of ten language-conditioned tasks, and a real \emph{SO-101} arm picks up a roll of tape (50 demonstrations). The
Panda benchmarks use global and wrist cameras with $H/E=16/8$ at 20\,Hz, while ALOHA uses $H/E=50/25$ at 50\,Hz.

For fair comparisons, every arm in a given test runs on the same reset seeds, time budget, and harness. Settings are
selected on development resets and frozen, and confirmations use fresh resets. A \emph{declared} test specifies its
hypothesis, endpoint, and multiple-testing correction before evaluation begins (Appendix~\ref{app:protocol}). Intervals
are 95\,\% bootstraps over resets of each arm's per-reset mean, and paired differences use a sign-flip test (resolution
$p\le10^{-4}$) with Holm correction within each declared family. Because absolute success varies across reset sets,
every claim rests on paired differences on shared resets. We call a difference \emph{separated} when its Holm-adjusted
$p<0.05$ (declared) or its interval excludes zero (descriptive); otherwise, it is not shown to be zero.

On ALOHA, we also track stricter handover endpoints because the standard benchmark reward peaks as soon as the left
gripper touches the cube off the table, before the handover finishes. Alongside success within 24\,s (and the
benchmark's 8\,s), we report \emph{strict success}, requiring the left gripper alone to hold the cube for 0.5\,s within
2\,s of success, and \emph{Held}, which continues the policy for 5--10\,s after success and verifies that the left
gripper still holds the cube (Appendix~\ref{app:endpoints}).

\subsection{Q1: What should the memory hand to the corrector?}
\label{sec:exp-mechanism}
\begin{figure}[t]
  \centering
  \includegraphics[width=\columnwidth]{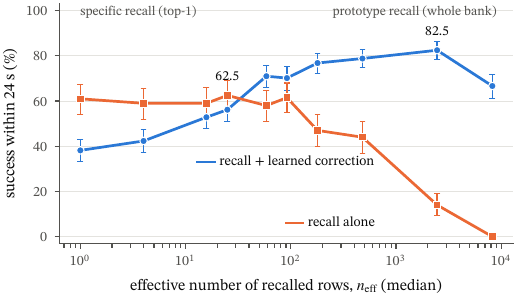}
  \caption{Readouts of the ALOHA Human memory with and without correction, plotted against the median number of rows
  averaged. Each corrected point averages three retrained heads (200 development resets; 95\,\% CIs).}
  \label{fig:reversal}
\end{figure}
\begin{figure*}[t]
  \centering
  \includegraphics[width=\textwidth]{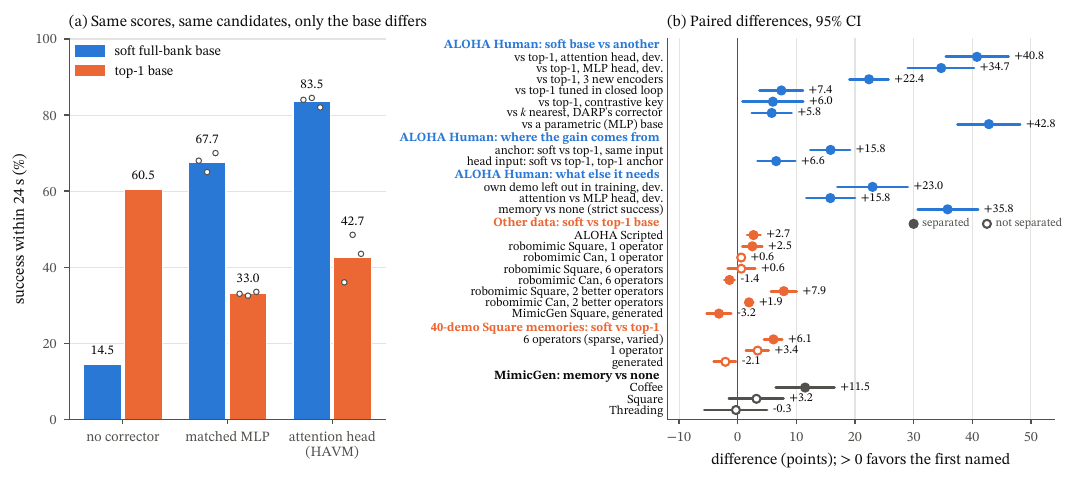}
  \caption{Soft base against top-1 base. (a) ALOHA Human on development resets: both bases come from the same
  $\beta=10$ scores and every corrector reads the same four candidates; dots denote three head seeds. (b) Paired
  differences in 24\,s success (unless marked) with 95\,\% intervals, grouped by comparison. Rows marked dev.\ use panel
  (a)'s resets, while others use fresh resets. ``Where the gain comes from'' crosses the anchor to which the correction is
  added with what the head reads (Table~\ref{tab:replication}'s encoders, heads, and resets). Filled markers are
  separated by their study's declared test. ``2 better operators'': trained on those alone.}
  \label{fig:mechanism}
\end{figure*}

We first sweep the ALOHA Human memory from sharp top-1 lookup to a broad softmax over thousands of rows on 200
development resets, training a separate corrector for each readout (Fig.~\ref{fig:reversal}). Without a corrector,
success stays around 58--62.5\,\% up to roughly $10^2$ effective rows (peaking at 62.5\,\% near 25 rows), but then
drops sharply once averaging pulls in too many distant, marginal memory rows, falling to 14.0\,\% near 2\,500 rows and
0\,\% near 8\,200 rows. With a corrector, the trend reverses: success starts low at top-1 (38.2\,\%), climbs steeply
across 25--100 rows to overtake uncorrected recall, and peaks at 82.5\,\% around 2\,500 effective rows before declining
to 66.7\,\% near the bank mean. Because the top-1 corrector in this sweep also sees fewer candidate rows,
Fig.~\ref{fig:mechanism}a isolates the base itself on the same development resets by giving both bases identical
$\beta=10$ similarity scores and the same four candidates. Without a corrector, top-1 succeeds in 60.5\,\% of trials
versus 14.5\,\% for the soft full-bank base. With the \ours{} corrector, their ordering flips to 42.7\,\% versus
83.5\,\% ($+40.8$\pp{} [35.5, 46.2]). To test whether this gain depends on our attention architecture,
Fig.~\ref{fig:mechanism}a also replaces the \ours{} corrector with a matched-size MLP (same parameter count and
dropout); the same reversal holds, with the soft base reaching 67.7\,\% versus 33.0\,\% for top-1 ($+34.7$\pp{}
[29.0, 40.3], both Holm $p\le10^{-4}$).

\begin{table}[t]
  \caption{Replication of the controlled comparison: three new encoder seeds (same recipe, each with its own
  validation choices), three heads per base, 200 fresh ALOHA Human resets, 24\,s success. TOP$^{\ast}$: top-1
  head with checkpoint and residual scale selected in closed loop on 20 development resets per setting. On the same
  resets, the development encoder (seed 0) with its heads reaches 88.3\,\%.}
  \label{tab:replication}
  \centering\footnotesize
  \setlength{\tabcolsep}{3.5pt}
  \begin{tabular}{@{}lccccc@{}}
    \toprule
    encoder & soft alone & top-1 alone & top-1 + head & TOP$^{\ast}$ & soft + head \\
    \midrule
    E1 & 66.0 & 62.0 & 49.7 & 65.5 & 75.0 \\
    E2 & 55.0 & 51.0 & 45.8 & 61.8 & 64.0 \\
    E3 & 36.5 & 53.0 & 45.7 & 58.7 & 69.3 \\
    \midrule
    mean & 52.5 & 55.3 & 47.1 & 62.0 & \textbf{69.4} \\
    \bottomrule
  \end{tabular}
\end{table}

Because those initial comparisons used a single development encoder on development resets, we ran additional
verification across three fresh encoder seeds and 200 fresh resets (Table~\ref{tab:replication}). Each new encoder
followed the same recipe with its own validation choices (all chose $\beta=40$, halved to a fixed 20 to hold $\beta$
constant across both bases before introducing offline $\beta$ learning in Section~\ref{sec:exp-main}), with three
correctors trained per encoder and base. With the corrector, the soft base leads by 22.4 points [19.1, 25.7]
($p\le10^{-4}$), and a bootstrap over encoders, correctors, and resets gives [13.5, 29.6]. Without correction, both
bases perform similarly at this sharper $\beta$ (52.5 vs.\ 55.3\,\%), so the gap comes from the corrector's effect:
$+16.9$ points on soft versus $-8.2$ on top-1. Corrector seeds vary more than encoder seeds (s.d.\ 5.6 vs.\ 4.0 points).
Because adding the default corrector lowered top-1 below top-1 alone (47.1 vs.\ 55.3\,\%) and our corrector recipe was
originally developed on the soft base, a natural concern is that the 22.4-point gap only reflects recipe bias or
over-correction on top-1. To rule this out, Table~\ref{tab:replication} includes a closed-loop-tuned top-1 control
(TOP$^{\ast}$) that selects its own checkpoint (out of six) and residual scale (0.1--1.0, six values) on 20 development
resets per setting; a 0.25 scale won on two encoders and the last checkpoint on the third. TOP$^{\ast}$ rises from
47.1 to 62.0\,\% (exceeding top-1 alone), confirming that much of the default top-1 drop came from over-correction. Even
with this closed-loop tuning advantage granted only to top-1, however, the untuned soft base (69.4\,\%) still leads
TOP$^{\ast}$ by 7.4 points [3.8, 11.1] (Holm $p=0.0003$), and by 7.2 [3.6, 10.7] against each encoder's best top-1
setting on test resets, including top-1 alone (without exclusion, a top-1 corrector learns nothing).

Next, we ask why the soft base helps: in~\eqref{eq:residual}, $A_{\mathrm{base}}$ acts both as the additive anchor
for the residual and as the query input read by the corrector. In a declared $2\times2$ study crossing the anchor and
input choices ($\{\text{top-1},\,\text{soft}\}\times\{\text{top-1},\,\text{soft}\}$) on Table~\ref{tab:replication}'s
encoders, correctors, and fresh resets (Fig.~\ref{fig:mechanism}b; Appendix~\ref{app:protocol}), anchoring on top-1
while reading soft recall gains 6.6 points [3.4, 9.8], whereas anchoring on soft recall adds 15.8 [12.4, 19.2]. A
soft-anchored corrector reading only top-1 still leads by 17.6 points. Roughly 70\,\% of the gain therefore comes from
using soft recall as the additive anchor, and in Held at 5\,s the input's share disappears.

Further ablations show that this gain depends on three conditions. First, the corrector must train on other
demonstrations: keeping the query's own demonstration in the bank drops success from 83.5 to 60.5\,\% ($+23.0$\pp{}
[17.0, 29.0]). Second, the key must reflect future actions. With a contrastive-only key (ALT's objective on our
encoder), soft recall fails alone (0\,\%), and corrected soft and top-1 bases reach 31.8 and 25.8\,\%, both below top-1
replay (39.2\,\%; that study's 200 resets). Soft recall remains easier to correct ($+6.0$\pp{} [0.8, 11.2]), but
correction does not pay off on an appearance-only key. Third, a parametric smooth base does not substitute: a
cross-fitted MLP on the same key, candidates, and corrector reaches 45.5\,\% versus 88.3\,\% for soft recall, though
alone it scores far higher (54.0 vs.\ 15.0\,\%). The same ordering also holds under an external corrector: in DARP's
mechanism~\cite{pfeiferDARP2026} on our key, soft-recall weighting of neighbor proposals outperforms $k$-NN averaging by
5.8 points [2.4, 9.2] (67.3 vs.\ 61.5\,\%, 400 fresh resets).

\begin{table*}[t]
  \caption{ALOHA cube transfer, 600 fresh resets per dataset. Baselines: LeRobot~\cite{cadeneLeRobot2024} recipes on
  \ours{}'s 40 human demonstrations, one training seed each (Appendix~\ref{app:baselines}). Reference rows in italic
  (excluded from bolding): the released ACT saw all 50 demonstrations, and the development encoder (seed 0) uses $\beta$
  20 / 80 (validation) and 10 / 40 (closed-loop tuned). \ours{} E1--E6: six fresh encoders $\times$ three heads, 18 runs
  per $\beta$ (learned $\beta^*$ 12.0--15.8; Scripted: development bank, validation 80, learned 25.6); ensemble: each
  encoder's three learned-$\beta$ heads averaged (Section~\ref{sec:head}), Scripted not run. Bold: best offline
  result with 40 demonstrations. Parameters exclude \ours{}'s bank (13\,M stored numbers, linear in demonstration count;
  59\,M on LIBERO). Training: wall time on one RTX\,5090. Policy call: p95, frame to chunk, dedicated GPU (\ours{}: with
  / without CUDA-graph replay); $^{\ddagger}$earlier session.}
  \label{tab:aloha}
  \centering
  \begin{tabular}{@{}lrrrrrrr@{}}
    \toprule
    & & & \multicolumn{3}{c}{Human} & Scripted & \\
    \cmidrule(lr){4-6}\cmidrule(lr){7-7}
    Policy & parameters & training & 24\,s & 8\,s & strict & 24\,s & policy call \\
    \midrule
    ACT~\cite{zhaoACT2023}                      & 52\,M & 2\,h\,03 & 71.7 & 59.5 & 65.3 & -- & 3.7\,ms \\
    Diffusion Policy~\cite{chiVisuomotorPolicyLearning} & 264\,M & 6\,h\,43 & 78.8 & 68.2 & \textbf{70.2} & -- & 364\,ms$^{\ddagger}$ \\
    \textit{ACT, released (50 demos)}           & \textit{52\,M} & -- & \textit{87.2} & \textit{71.8} & \textit{80.5} & -- & \textit{3.6\,ms} \\
    X-VLA~\cite{zhengXVLA2025}, fine-tuned      & 880\,M & 1\,h\,28 & 75.2 & 61.0 & 68.0 & -- & 44\,ms \\
    SmolVLA~\cite{shukorSmolVLA2025}, fine-tuned & 450\,M & 2\,h\,32 & 65.2 & 62.5 & 60.8 & -- & 86\,ms \\
    \midrule
    Soft recall alone, validation $\beta$        & 11\,M & 2\,min & 46.7 & 46.0 & 39.0 & 62.3 & 1.5\,ms \\
    Top-1 + MLP residual (ReCAP-style)          & 14\,M & 3\,min & 51.8 & 51.0 & 39.0 & \textbf{97.3} & 2.7\,ms \\
    \ours{}, E1--E6, validation / learned $\beta$ (offline) & 14\,M & 3 / 5\,min & 72.7 / 76.4 & 71.5 / 74.6 & 60.8 / 64.0 & 96.2 / 96.2 & 1.1 / 2.6\,ms \\
    \ours{} + 3-head ensemble, E1--E6, learned $\beta$ & 20\,M & 8\,min & \textbf{79.5} & \textbf{77.6} & 69.1 & -- & 1.5 / 2.8\,ms \\
    \textit{\ours{}, dev.\ encoder, valid.\ / closed-loop $\beta$} & \textit{14\,M} & \textit{3\,min} & \textit{77.8 / 82.9} & \textit{76.5 / 81.8} & \textit{64.9 / 74.4} & \textit{96.2 / 97.8} & \textit{1.1 / 2.6\,ms} \\
    \bottomrule
  \end{tabular}
\end{table*}
To test how broadly this behavior generalizes beyond ALOHA Human, we repeat the soft-versus-top-1 comparison across
other datasets (Fig.~\ref{fig:mechanism}b, fresh resets). The corrected soft base leads by 2.7 points on scripted ALOHA
and 2.5--2.9 on single-operator robomimic Square, matches top-1 on full six-operator Square (51.6 vs.\ 51.0\,\%), and
trails by 1.4 on full six-operator Can and 3.2 on generated MimicGen Square. These results also show that \ours{}
depends on demonstration quality: trained on just 90 demonstrations from the two better operators (by skill labels), it
reaches 63.2\,\% on Square ($+7.9$ over top-1's 55.3\,\%) and 99.5\,\% on Can ($+1.9$), substantially outperforming a
size-matched 90-demonstration draw across all six mixed-skill operators (32.2 vs.\ 25.5\,\%, $+6.7$ on Square; 91.5
vs.\ 89.1\,\%, $+2.4$ on Can) and even exceeding the full 270-demonstration mixed-skill set on Square (51.6\,\%). Mixing
in low-skill demonstrations degrades the averaged base and lowers overall success, whereas higher-quality human subsets
preserve both strong accuracy and a clear soft-base gain. Our first declared prediction, that full multi-operator data
would yield a larger soft-over-top-1 gain than single-operator data, failed (Square $-2.3$, Can $-2.0$\pp{}, Holm
$p=0.0007$) because of mixed operator quality and higher demonstration density. Since ALOHA Human's grasps are sparser
(1.39\,cm between neighbors vs.\ 0.50--0.57\,cm on full robomimic and MimicGen Square), we then declared the relative
gain would be largest on sparse multi-operator human data. On 40-demonstration Square
memories (1\,000 fresh scenes), multi-operator memory indeed gains 6.1 points [4.5, 7.6], exceeding the other three
40-demonstration settings (all Holm-separated), though absolute success stays low (8--24\,\%) due to sparse coverage
and mixed skill. Post hoc across all fourteen memories, gains above 5 points occur only on human memories of 40--90
demonstrations (Appendix~\ref{app:rule}), matching the controlled toy (Appendix~\ref{app:toy}).

Finally, we examine how sensitive the corrector is to the recall bandwidth $\beta$. Selecting $\beta$ by the validation
chunk error of uncorrected recall yields 20 on Human and 80 on Scripted, which is too sharp for the corrector. Halving
each value improves closed-loop success on 600 fresh resets ($+5.1$\pp{} [2.8, 7.4] on Human and $+1.6$\pp{} [0.9, 2.4]
on Scripted); Human recall then averages roughly 2\,450 rows versus 53 on Scripted, because human operators disagree on
grasp timing (Appendix~\ref{app:beta}). Sections~\ref{sec:exp-main} and~\ref{sec:exp-when} test
replacing this closed-loop halving with offline $\beta$ learning.

\subsection{Q2: Against parametric and retrieval-based policies}
\label{sec:exp-main}
\begin{figure}[t]
  \centering
  \includegraphics[width=\columnwidth]{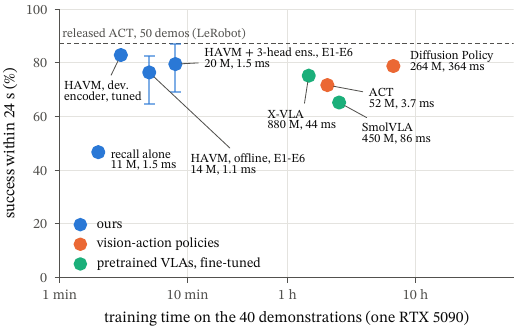}
  \caption{Success on ALOHA Human against training time on the same 40 demonstrations (labels: parameters, policy
  call p95), every policy on the 600 resets of Table~\ref{tab:aloha}. \ours{}: E1--E6 with offline learned $\beta$ for
  one head (18 runs) and the fused three-head ensemble (6 encoders; bars show the min--max range across the six
  encoders), plus the development encoder at its tuned $\beta$.}
  \label{fig:cost}
\end{figure}

We first compare \ours{} against parametric vision-action policies trained on the same 40 ALOHA demonstrations
(Table~\ref{tab:aloha}, Fig.~\ref{fig:cost}). Because the LeRobot baselines evaluate their final checkpoint without
closed-loop selection, the fair comparison uses \ours{} with offline choices across six fresh encoders (E1--E6, three
correctors each). With learned $\beta$ (Section~\ref{sec:execution}), \ours{} reaches 76.4\,\% at 24\,s, matching
Diffusion Policy ($-2.4$\pp{} [$-5.9$, 1.0], Holm $p=0.17$), outperforming ACT ($+4.7$\pp{} [1.1, 8.5], $p=0.029$) and
fine-tuned VLAs (65.2--75.2\,\%), and leading every 40-demonstration baseline within the 8\,s limit (74.6\,\%). As
Fig.~\ref{fig:cost} shows, \ours{} sits in the upper-left corner of the accuracy--cost trade-off: it trains in five
minutes with 14\,M trained parameters (plus the memory bank) and runs in 1.1\,ms per call, one to two orders of
magnitude faster to train than ACT, Diffusion Policy, and the fine-tuned VLAs (52\,M--880\,M parameters, 1.5--6.7\,h
training, 3.7--364\,ms calls). Two nuances remain: \ours{} trails Diffusion Policy in strict handover success
($-6.1$\pp{} [$-10.0$, $-2.2$]), and encoder-seed spread (64.6--82.7\,\% with learned $\beta$ across E1--E6, shown by the
range bar in Fig.~\ref{fig:cost}; 62.2--83.4\,\% at validation $\beta$) outweighs the choice of $\beta$ (learned $\beta$
adds 3.8 points overall, and bootstrapping over encoders, correctors, and resets includes zero against both ACT and
Diffusion Policy). When trained on all 50 demonstrations (matching the released ACT on 600 fresh resets), \ours{}
reaches 83.2\,\% versus ACT's 88.2\,\%, rising to 92.1\,\% with the post-hoc absolute-error loss
(Section~\ref{sec:exp-when}).

Averaging each encoder's three learned-$\beta$ heads (Section~\ref{sec:head}) closes most of the strict-success gap,
adding 5.0 points [1.4, 8.8] in strict success over encoders and resets (improving 5 of 6 encoders) and 3.1 points
[2.1, 4.2] at 24\,s over resets ([1.2, 4.9] over encoders; 6 of 6). On Table~\ref{tab:aloha}'s resets, this reaches
79.5\,\% at 24\,s (69.2--87.0\,\% across E1--E6 in Fig.~\ref{fig:cost}), 77.6\,\% at 8\,s, and 69.1\,\% strict,
comparable to Diffusion Policy (78.8\,\% at 24\,s and 70.2\,\% strict) while raising training time only from 5 to 8
minutes (20\,M parameters, 1.5\,ms per call).

\begin{table}[t]
  \caption{DARP's protocol and code (MimicGen D0, first 100 demonstrations, its simulation stack): success on the
  same 200 trials, mean of three seeds, all our runs (published: 52/75/76\,\% after 65--322 tuning runs). Features:
  R3M at 224\,px for DARP, ResNet-18 at 128\,px for \ours{}. Correction vision (Threading only): 8.5\,M extra trainable
  vision parameters, $K_c=0$. Bold: highest per task; on Threading, 62.3 and 57.5 are not separated.}
  \label{tab:darp}
  \centering\small
  \begin{tabular}{@{}lccc@{}}
    \toprule
    Policy & Square & Stack & Threading \\
    \midrule
    BC (DARP's code, R3M)                   & 10.2 & 15.5 & 19.7 \\
    DARP~\cite{pfeiferDARP2026}             & 34.5 & 42.8 & 51.0 \\
    DARP, distance-weighted                 & 36.2 & 38.7 & 51.0 \\
    DARP, other-demo neighbors              & 34.0 & \textbf{44.0} & \textbf{57.5} \\
    \midrule
    \ours{}, valid.\ $\beta$                & 39.2 & 35.5 & 19.5 \\
    \ours{} ($\beta=1$)                     & \textbf{44.8} & 42.8 & 36.7 \\
    \ours{} + correction vision             & -- & -- & \textbf{62.3} \\
    \bottomrule
  \end{tabular}
\end{table}

We next compare against retrieval-based policies, starting with DARP~\cite{pfeiferDARP2026} under its official MimicGen
protocol and code (Table~\ref{tab:darp}). \ours{} ($\beta=1$) leads DARP on Square ($+10.3$\pp{} [4.8, 15.8]), matches
it on Stack, and trails on Threading ($-14.3$\pp{}), where frozen retrieval features miss fine alignment; adding
trainable image features to the corrector (correction vision) raises Threading to 62.3\,\%, beating standard DARP
($+11.3$\pp{} [5.8, 16.8]) and matching its other-demonstration variant. Note that our local runs of DARP's released
code score below its published figures (34/40/53\,\% vs.\ 52/75/76\,\% on tuning trials), which \ours{} also trails.

Finally, across eight retrieval baselines re-implemented under a unified ALOHA Human harness with equal closed-loop
tuning budgets (MCNN, REGENT, DARP, BPC, VINN, ALT, retrieve-and-play, and ReCAP's top-1 residual;
Appendix~\ref{app:retrieval}), \ours{} outperforms every alternative by a wide margin: 85.0\,\% versus 51.5\,\% for
top-1 residual and 50.0\,\% for MCNN (200 resets, Holm $p\le10^{-4}$), and 85.4\,\% versus 61.5\,\% for the strongest
external baseline, DARP's mechanism on \ours{}'s key (400 resets). These adaptations omit system-specific components
such as REGENT's pretraining and are evaluated on \ours{}'s development benchmark.

\subsection{Q3: Many tasks, new wording, a real robot, and more data}
\label{sec:exp-breadth}
\begin{table}[t]
  \caption{LIBERO with one 40-task memory (top: three seeds $\times$ 500 episodes per suite; gray: published, other
  protocols~\cite{shukorSmolVLA2025,kimOpenVLAOFT2025}, $\pi_0$~\cite{blackPi02024} as reported in~\cite{kimOpenVLAOFT2025}) and under new wording and moved objects (bottom:
  original/changed on the same episodes, seed 0; rewording 10 initial states per task, 400 episodes; moves 20,
  800). $^{*}$Darker than the demonstrations here.}
  \label{tab:libero}
  \centering\footnotesize
  \setlength{\tabcolsep}{2.2pt}
  \begin{tabular}{@{}lccccc@{}}
    \toprule
    & Spatial & Object$^{*}$ & Goal & Long & mean \\
    \midrule
    Soft recall alone          & 92.0 & 99.1 & 95.9 & 81.3 & 92.1 \\
    \ours{}                    & 95.7 & 97.5 & 93.9 & 81.8 & 92.2 \\
    {\color{gray}SmolVLA (0.45\,B)}  & {\color{gray}90} & {\color{gray}96} & {\color{gray}92} & {\color{gray}71} & {\color{gray}87.3} \\
    {\color{gray}$\pi_0$, fine-tuned} & {\color{gray}96.8} & {\color{gray}98.8} & {\color{gray}95.8} & {\color{gray}85.2} & {\color{gray}94.2} \\
    {\color{gray}OpenVLA-OFT}  & {\color{gray}97.6} & {\color{gray}98.4} & {\color{gray}97.9} & {\color{gray}94.5} & {\color{gray}97.1} \\
    \midrule
    \multicolumn{6}{@{}l}{\emph{verb substitutions, original / rewritten}} \\
    \ours{}                    & 98/96 & 100/90 & 94/91 & 79/72 & 92.8/87.3 \\
    SmolVLA                    & 82/46 & 93/57 & 90/56 & 68/62 & 83.3/55.3 \\
    \multicolumn{6}{@{}l}{\emph{objects moved 6\,cm, unmoved / moved}} \\
    \ours{}                    & 97/60 & 99.5/50.5 & 95.5/66.5 & 79.5/37 & 92.9/53.5 \\
    SmolVLA                    & 76.5/31 & 84/31.5 & 83.5/56.5 & 72/24.5 & 79.0/35.9 \\
    \bottomrule
  \end{tabular}
\end{table}

To test multi-task scaling and language robustness, we store all four LIBERO suites in a single memory (40
instructions, 1\,533 demonstrations, 246\,669 rows), which reaches 92.2\,\% at 8\,ms per call (Table~\ref{tab:libero}).
Soft recall alone already scores 92.1\,\%, open-loop replay of the nearest demonstration succeeds in 70\,\% of episodes,
and shifting every object by 3 or 6\,cm lowers success to 80 and 54\,\%, confirming that standard LIBERO primarily tests
recall~\cite{zhouLIBEROPRO2025,feiLIBEROPlus2025}. We therefore test robustness against the authors' fine-tuned SmolVLA
on \ours{}'s episodes after passing a pre-set validity threshold (Appendix~\ref{app:vla}). Rewordings fixed before
\ours{}'s runs degrade the VLA much more severely: verb substitutions cost \ours{} 5.5 points versus 28.0 for SmolVLA
(difference $-22.5$\pp{} [$-28.7$, $-16.0$]), while Goal sentence rewrites cost 25 and 58 points ($-33.0$\pp{} [$-49$,
$-17$], Holm $p=0.0004$). Moving objects by 6\,cm degrades both policies similarly ($-3.8$\pp{} [$-9.0$, 1.6]).

Beyond simulation, we evaluated \ours{} on a real SO-101 arm in a tape-picking task (45 demonstrations). In 10 blind
paired trials, \ours{} succeeded 7 times and a fine-tuned SmolVLA 8 times ($p=1.0$). SmolVLA's 81\,ms re-plan stalls the
33\,ms control loop, whereas \ours{}'s 6\,ms re-plan does not (Appendix~\ref{app:robot}).

Finally, we check how \ours{} scales across additional datasets and demonstration counts. On robomimic, the same recipe
reaches 99.0\,\% on Can and 83.7\,\% on Square, close to BC-RNN's published results~\cite{mandlekarRobomimic2021}. On
ALOHA Human, performance jumps sharply between 15 and 20 demonstrations; retraining on all 50 adds 1.8 points (not
separated), while appending ten demonstrations to a frozen memory does not help ($-3.0$\pp{} [$-6.0$, 0.0];
Appendix~\ref{app:data}).

\subsection{Q4: When do design choices matter?}
\label{sec:exp-when}
\begin{table}[t]
  \caption{Three design choices across three data sources: paired differences in points (95\,\% CI) between otherwise
  identical arms, three seeds each, fresh resets. Closure spread: s.d.\ of first gripper closure across demonstrations.
  $^{a}$Strict success. $^{b}$Held 5--10\,s after success, outer quarter of cube starts. $^{c}$Benchmark success; Square,
  Threading, Coffee in order. $^{d}$Square, 200 generated demonstrations. $^{e}$Table~\ref{tab:replication}'s $3\times3$
  runs.}
  \label{tab:when}
  \centering\scriptsize
  \setlength{\tabcolsep}{1.4pt}
  \begin{tabular}{@{}lccc@{}}
    \toprule
    & ALOHA Human & ALOHA Scripted & MimicGen \\
    \midrule
    demonstrations & 40 human & 40 scripted & 100 generated \\
    closure spread & 0.62\,s & 0 & 0.34--0.57\,s \\
    \midrule
    soft $-$ top-1 & $+22.4^{e}$ [19.1, 25.7] & $+2.7$ [1.6, 3.9] & $-3.2^{d}$ \\
    memory $-$ none & $+35.8^{a}$ [30.7, 41.0] & -- & $+3.2$, $-0.3$, $+11.5^{c}$ \\
    abs.\ $-$ sq.\ loss$^{b}$ & $+12.2$ [6.4, 18.0] & $-4.9$ [$-8.0$, $-2.0$] & --, $-1.2$, $-5.5^{c}$ \\
    \bottomrule
  \end{tabular}
\end{table}

Design choices that help on human demonstrations often have little effect, or reverse, on scripted and generated data
(Table~\ref{tab:when}). Human operators close the gripper at varying times, causing soft recall to smear closure over
37--72 steps versus 13 in the demonstrations. Training the corrector with absolute instead of squared error keeps
closure sharper and raises Held in the outer quarter of cube starts by 12.2 points (Holm $p=0.0005$, 600 fresh resets).
On Scripted, where every demonstration closes at the exact same step, absolute error costs 4.9 points, and on MimicGen
it adds no gain, so squared error remains the default.

A similar sensitivity appears in hyperparameter selection: offline chunk error of uncorrected recall repeatedly
misranked settings, so keeping the baseline offline defaults (validation $\beta$, soft base, squared error) costs 5.1
points on ALOHA Human (development encoder), 1.6 on Scripted, and 8.5--13.2 on MimicGen, with no measurable loss on
LIBERO or robomimic. Learning $\beta$ offline through the leave-one-demonstration-out corrector loss closes most of this
gap on ALOHA Human (landing at 11--16 versus a tuned 10 and adding 3.8 points across E1--E6; Section~\ref{sec:exp-main})
and moves robomimic Square from 20 to 10.9 (not run in closed loop). Elsewhere it does not replace closed-loop
selection: on Scripted it stops at 25.6 versus a tuned 40 and matches the validation $\beta$ (96.2\,\%), 1.6 points
below the closed-loop 97.8\,\%, and on MimicGen Square and Threading, where a corrector without memory performs just as
well, it stops at 28.6 and 16.4 against a tuned 1--5, failing its declared cross-benchmark check
(Appendix~\ref{app:protocol}). Validation under input drift ranks $\beta=1$ highest on ALOHA, which is its worst
closed-loop setting. We recommend the soft base for small human datasets and top-1 for generated or plentiful
demonstrations.

Corrector seed variance on the same frozen encoder also matters. On ALOHA Human, sibling heads match in validation error
to about 0.01$^\circ$, yet their corrections on identical frames differ by 57--71\,\% of their RMS and their closed-loop
success swings by up to 15--20 points. More conservative single heads (shorter training, stronger regularization, or a
narrower network) and longer training (epochs 20--60) both lowered strict success by 4.2--12.1 points
(Appendix~\ref{app:negative}). Within an encoder, strict success rises with correction magnitude ($\rho=+0.73$ over 36
heads, post hoc; 4.7$^\circ$ per step on the right arm), because grasping needs a strong correction against the soft
base's temporal smear, while longer training adds seed-specific error. Averaging three heads keeps the shared correction
and cancels much of that seed error. Across benchmarks, the gain grew with within-encoder head disagreement (s.d.):
Square PH $+1.7$ [$-0.1$, 3.5] (s.d.\ 2.7; 5 encoders better, none worse), ALOHA $+5.0$ strict (s.d.\ 3.5; $+6.2$
[3.6, 8.7] on 12 further encoders), and Threading $+8.8$ [2.5, 16.0] (s.d.\ 5.0; 6 of 6). On Threading's strongest
variant (correction vision, $K_c=0$; 200 fresh trials), it raises success from 61.3 to 68.3\,\% ($+7.0$ [2.9, 11.1];
per encoder $+0.5$ to $+12.7$, reaching 79.5\,\%). The ensemble matched or beat its worst single head on all 33
encoders. An earlier test gave $+7.7$ [3.3, 12.0] on MimicGen Square but $-2.2$ [$-6.2$, 1.8] strict on ALOHA's
development encoder. Ensembling removes head-level blow-ups, while encoder-level variance did not shrink on any
benchmark.

\subsection{Q5: Limitations and contrasting strengths}
\label{sec:exp-limitations}
A primary structural limitation of \ours{} is that both the softmax base $A_{\mathrm{base}}=\sum_i w_iV_i$ and the
deterministic regression corrector $r_\phi$ predict one action chunk instead of a distribution over distinct action
modes. Averaging across demonstrations works well on relatively simple tasks where demonstrations follow one shared
strategy with operator jitter in timing or trajectory. On genuinely multi-modal tasks where distinct paths are equally
valid from the same observation (e.g., passing an obstacle to the left or the right), soft recall and squared-error
regression can cancel opposing modes and steer into an invalid intermediate trajectory, whereas diffusion and
flow-matching policies~\cite{chiVisuomotorPolicyLearning,lipmanFlowMatching2023} model multiple modes directly. Even on
our unimodal human tasks, averaging smears gripper closure over 37--72 steps versus 13 in demonstrations
(Section~\ref{sec:exp-when}), requiring the corrector to sharpen it back. By contrast, when a single strategy suffices,
this deterministic single-pass design avoids iterative denoising and sampling variance, cutting ALOHA inference from
364\,ms (Diffusion Policy) and 44--86\,ms (fine-tuned VLAs) to 1.1\,ms (1.5\,ms for three heads) and achieving the
highest 8\,s completion rate (74.6\,\% with one head and 77.6\,\% with three, Table~\ref{tab:aloha}).

A second limitation concerns contact precision and coverage beyond the stored demonstrations. Empirically, failures
concentrate at first contact (missed grasps on ALOHA, unlifted nuts on Square, the needle missing the ring, and fingers
hitting the tape) and track demonstration coverage closely. In the outer quarter of ALOHA's cube starts, \ours{}
completes 57\,\% of handovers versus 89--95\,\% in the interior, and outside the demonstrated 20$\times$20\,cm square it
succeeds at only 2 of 338 positions (released ACT: 4); neither extrapolating corrections nor thinned memories improved
edge performance (Appendices~\ref{app:negative},~\ref{app:thin}). Likewise, on LIBERO \ours{} cannot follow genuinely
novel instructions outside its 40 stored tasks, whereas pretrained VLAs attempt open-vocabulary transfer
(Section~\ref{sec:exp-breadth}). Within its demonstrated workspace and task set, however, explicit retrieval gives
transparent, inspectable predictions (each base traces directly to weighted stored demonstrations), trains in about five
minutes with 14\,M parameters compared with hours on 52\,M--880\,M parameters, and degrades far less than fine-tuned
SmolVLA under local instruction rewording ($-5.5$ vs.\ $-28.0$ points on verb substitutions, Table~\ref{tab:libero}).

Finally, \ours{}'s advantages taper outside the low-data regime, while memory and tuning costs grow with task
diversity. When demonstrations are dense, scripted, or abundant, the soft-base gain fades or slightly reverses
(Section~\ref{sec:exp-mechanism}), a memory-free corrector matches \ours{} on MimicGen Square and Threading
(Table~\ref{tab:when}), and a large behavior-cloning MLP overtakes state-keyed \ours{} on 1\,000 scripted OCBench
demonstrations (Appendix~\ref{app:ocbench}). Bank storage and similarity lookup also scale linearly with stored frames
(from 16\,000 rows / 1.1\,ms on ALOHA to 246\,669 rows / 8\,ms on LIBERO), appending demonstrations to a frozen memory
does not improve performance without retraining (Section~\ref{sec:exp-breadth}), and while three-head ensembling
removes head-seed blow-ups, results remain sensitive to encoder seed (62--83\,\% across E1--E6) and to $\beta$ selection
where offline learning does not transfer across benchmarks (Section~\ref{sec:exp-when}). Conversely, in sparse-data, compute-constrained, or real-time settings (such as
40-demonstration ALOHA or the 30\,Hz SO-101 arm, where SmolVLA's 81\,ms re-plan stalls the 33\,ms control loop while
\ours{} runs in 6\,ms), \ours{} matches larger policies at a fraction of their parameters, training time, and latency.

\section{Discussion and Limitations}
\label{sec:discussion}
Top-1 recall commits to one demonstration and switches abruptly as the robot moves, so its error stays small when it
picks the right trajectory but jumps sharply when it does not. Soft recall averages many demonstrations: biased toward
their mean and poor alone when they disagree, its error varies smoothly with the observation and is easier for a
conditioned corrector to fit. The soft base is a high-bias, low-variance smoother whose residual the head models, akin
to twicing~\cite{tukeyEDA1977,buhlmannL2Boost2003}, while pairing a fast associative store with a slowly trained network
resembles complementary learning systems~\cite{mcclellandCLS1995,kumaranCLS2016}. On held-out demonstrations, soft
recall's error changes half as fast relative to its magnitude as top-1's (Appendix~\ref{app:why}), and most of its
benefit comes from anchoring the correction (Section~\ref{sec:exp-mechanism}).

A state-keyed toy (Appendix~\ref{app:toy}; four of seven declared predictions held) clarifies when this soft-base gain
appears: soft recall never helps when demonstrations agree, helps on average when they disagree and are sparse (one cell
$-3.3$, n.s.), including with six mixed-skill operators, and fades at 270 demonstrations ($+6.1$, n.s.). Across fourteen
real memories, demonstration count orders the gain while no offline statistic does; sparse data likely leave the head
unable to learn top-1's jumps. How much the memory itself contributes also varies by task: recall carries ALOHA Human,
MimicGen Coffee, and LIBERO (92\,\% alone), whereas on MimicGen Square and Threading a head without memory performs just
as well.

Several caveats remain in our evaluation. ALOHA Human was our development benchmark. Encoders from one recipe span
62--83\,\% at validation $\beta$ (an 11-point gap between E1--E3 and E4--E6 means). Initialization and minibatch order
with augmentation contribute about equally (s.d.\ 4.0 and 4.6 points on nine crossed encoders, roughly additive), and
weak encoders lose mainly initial grasps decided during the approach. Mean key cosine, which had separated encoders post
hoc, reversed sign on those nine encoders (Appendix~\ref{app:negative}). While the head ensemble removes head-seed
blow-ups, reducing encoder spread remains open.
Finally, \ours{} does not extrapolate beyond its demonstrated workspace, a large behavior-cloning network overtakes
state-keyed \ours{} on abundant scripted state data (Appendix~\ref{app:ocbench}), and learned $\beta$ misses
MimicGen's soft optimum, leaving closed-loop tuning costly on robots.

\section{Conclusion}
\label{sec:conclusion}
A demonstration memory should supply its corrector with a smooth recall when demonstrations are sparse and varied,
whereas on dense or generated data the nearest neighbor performs equally well or slightly better. On ALOHA's human
demonstrations and 90-demonstration robomimic Square memories, correcting soft recall outperforms correcting the nearest
chunk, primarily because it provides a better anchor under a future-action key and leave-one-demonstration-out training.

\bibliographystyle{IEEEtran}
\bibliography{IEEEabrv,references}

\clearpage
\appendix

\subsection{Configurations, data splits, and training}
\label{app:config}
\begin{table}[ht]
\caption{Configurations. $H/E$: chunk length / executed prefix; squared-error loss and a validation-chosen head
unless marked. $\beta$: closed-loop or validation choice; $^{\dagger}$or the learned $\beta$, ALOHA only ($\beta^*$:
10.9 dev.\ encoder, 12.0--15.8 E1--E6, 25.6 Scripted).}
\label{tab:config}
\centering\footnotesize
\setlength{\tabcolsep}{2.5pt}
\begin{tabular}{@{}L{2.0cm}L{2.35cm}cL{1.05cm}L{1.35cm}@{}}
\toprule
Policy & Bank, cameras & $H/E$ & $\beta$, $K_c$ & Head \\
\midrule
ALOHA Human & 40 demos, top & 50/25 & 10$^{\dagger}$, 4 & validation \\
ALOHA Scripted & 40 demos, top & 50/25 & 40$^{\dagger}$, 4 & validation \\
ALOHA Human, all 50 & 50 demos, 40-demo encoder & 50/25 & 10, 4 & abs.\ or sq.; epoch 10 fixed \\
robomimic Can, Square & 180 demos, global + wrist & 16/8 & 20, 4 & validation \\
MimicGen (DARP protocol) & 90 demos, global + wrist & 16/8 & 1, 4 & validation \\
Threading + correction vision & as above, trainable last ResNet stage & 16/8 & 1, 0 & validation \\
MimicGen scaling & 100--950 demos, global + wrist & 16/8 & valid., 4 & validation \\
LIBERO, 40 tasks & 1\,533 demos (246\,669 rows), global + wrist, language & 16/8 & 20, 4 & validation \\
SO-101 & 45 demos, top + wrist & 30/15 & valid., 4 & validation \\
\bottomrule
\end{tabular}
\end{table}
The ALOHA bank uses demonstrations 0--39, with 40--44 selecting the head. In the all-50 run, the 40-demonstration encoder
is kept, all 50 demonstrations enter the bank, and the epoch-10 EMA head is fixed beforehand because the validation
demonstrations now sit in the bank. robomimic uses 180 training and 20 validation demonstrations, the 100-demonstration
MimicGen pool splits 90/10, and LIBERO holds out the last four demonstrations per instruction for validation
(Table~\ref{tab:config}). Frames enter at 168$\times$224\,px on ALOHA and 128$\times$128\,px elsewhere. The encoder
(ImageNet ResNet-18, 128-d keys) trains for 20 epochs with AdamW (lr $3\cdot10^{-4}$, weight decay $10^{-4}$, batch 256)
on two views per frame (random resized crop 70--100\,\%, color jitter 0.4 with probability 0.8, state noise 0.05) under
NT-Xent (temperature 0.4) plus the action-aware loss. Every two epochs, the validation chunk error of recall ($\beta=40$)
scores the encoder and selects the best checkpoint; $\beta$ is then chosen by the same error. The binding weight is
$\lambda_\ell=1$. Multi-task instructions are encoded by SmolVLM2-500M-Video-Instruct's language model (layer 16,
mean-pooled over at most 48 tokens). The head trains for 60 epochs with AdamW (lr $3\cdot10^{-4}$, weight decay $10^{-6}$,
batch 256, OneCycle with 5\,\% warm-up, gradient-norm clipping 1, EMA decay 0.995). On a dedicated GPU, training takes
about 100\,s for an ALOHA encoder, 70\,s for a head, and 30 minutes on the 246\,669-row LIBERO memory. Learned $\beta$
(ALOHA only) trains $\log\beta$ (lr $3\cdot10^{-2}$, no weight decay, separate gradient-norm clip 1) jointly with a head
for 60 epochs from validation $\beta$ ($\sim$100\,s); its epoch-60 EMA value $\beta^*$ is kept and a fresh head (70\,s)
trains at $\beta^*$. Candidate weights share the base's $\beta$. Base tokens carry a constant embedding inherited from
the velocity network whose code the head reuses; nothing is integrated. For Threading, the visual variant (8.5\,M extra
parameters) copies ResNet layer 4 and its projection into a branch trained by the action loss at one-tenth of the head's
learning rate with frozen BatchNorm statistics, while the retrieval encoder and bank remain frozen. All runs used one RTX
5090. Fig.~\ref{fig:mechanism}a reuses the ALOHA resets (8\,100\,000--8\,100\,199) and robomimic scenes
(8\,200\,000--8\,200\,199) from the readout sweep with shared soft-base references; Fig.~\ref{fig:mechanism}b uses fresh
resets.

\subsection{Declared tests}
\label{app:protocol}
Each declared test in Table~\ref{tab:protocol} fixed its hypothesis, endpoint, reset set, and multiple-testing
correction before evaluation began.
\begin{table}[ht]
\caption{Declared tests and outcomes (endpoint: 24\,s success on ALOHA, success elsewhere; fresh resets; unless
marked). Held and failed refer to declared directions. In robomimic, PH denotes proficient-human (single operator) and
MH denotes multi-human (six operators of mixed skill).}
\label{tab:protocol}
\centering\scriptsize
\setlength{\tabcolsep}{2pt}
\begin{tabular}{@{}L{4.25cm}L{4.15cm}@{}}
\toprule
Test (resets) & Outcome \\
\midrule
$\beta\times\frac12$ against $\beta$ (600 each) & held: $+5.1$, $+1.6$ \\
Soft against top-1, two correctors (200 dev.) & held: $+40.8$, $+34.7$ \\
Mechanisms under equal tuning (200) & held: 85.0 against at most 51.5\,\% \\
DARP's code, its protocol (200 trials) & Square $+10.3$, Stack 0, Threading $-14.3$ \\
Soft base, Square (1\,000), Scripted (600) & $+2.5$, $+2.7$, both separated \\
MH gain above PH gain, robomimic (1\,000) & \textbf{failed}: Square $-2.3$, Can $-2.0$ \\
MimicGen; MH above PH; PH above MimicGen & $-3.2$; \textbf{failed} ($-0.9$); held ($+6.1$) \\
Soft weighting in DARP's corrector (400) & held: $+5.8$ \\
Square-40 memories, first test (200) & $+6.2$, under the predicted 10 \\
The rule on Square-40 memories (1\,000) & held: all three contrasts \\
Absolute-error head (600; Held) & held: $+12.2$ at the edge \\
SmolVLA, rewording and shifts (paired) & rewrites $-33.0$; shift $-3.8$, n.s. \\
Contrastive key (200, reused) & soft over top-1 $+6.0$; both below replay \\
Replication; tuned top-1; parametric; growth & $+22.4$; $+7.4$; 88.3 vs.\ 45.5\,\%; $-3.0$ \\
Thinned memories, pilot (200 each; Held) & no rate changed Held \\
Finer top-1 scales; anchor vs.\ input (reused) & unchanged; anchor $+15.8$, input $+6.6$ \\
Better operators only, MH (1\,000, reused) & held: $+7.9$ \\
Toy (500 per head); 14-memory measures & 4 of 7 held; 1 of 3 held \\
E1--E3 (valid.\ $\beta$) vs.\ DP, ACT (600) & $-11.7$, $-4.6$, separated \\
Learned $\beta$ (200); drifted val. & ALOHA held; benchmark gate \textbf{failed} (MimicGen); drifted \textbf{failed} \\
Learned $\beta$ vs.\ DP, ACT: E4--E6; E1--E6; vs.\ valid.\ $\beta$ (600) & $-0.3$, $+6.9$; $-2.4$ (level), $+4.7$; $+3.8$ \\
Key uniformity: ALOHA (600 reused); Threading; Square & \textbf{failed}: $+2.6$; $-0.1$; $-3.7$ \\
$E=5$ + temporal ensembling (600 reused; 200; 200) & $+2.0$; strict $+7.0$; strict $+1.7$, n.s. \\
Paired crop; with fixed epoch 12 (200) & \textbf{failed}; \textbf{failed} (seed s.d.) \\
Conservative heads; late epochs (200; strict) & \textbf{failed}: $-4.2$ to $-12.1$; $-6.3$ to $-7.8$ \\
Head ensemble: 12 enc.; E1--E6 (reused); Square; Threading; Thr.\ + vision & held: strict $+6.2$; strict $+5.0$; $+1.7$; $+8.8$; $+7.0$ \\
Ensemble call without graph (harness) & 2.74 vs.\ 2.70\,ms (one head), ACT 3.53 \\
\bottomrule
\end{tabular}
\end{table}

\subsection{ALOHA endpoints}
\label{app:endpoints}
A gripper \emph{holds} the cube when both of its fingers have touched it within the last ten steps, the cube does
not touch the table, and its center is more than 1\,cm above its resting height. After the first benchmark success,
\emph{A3} requires 25 consecutive left-held steps (0.5\,s at 50\,Hz) within the next 100 steps, without a return to
the table, and \emph{A3s} also requires no right-finger contact in the ten steps before each of them.
\emph{Strict success} is A3s, or a left hold through the last 25 recorded steps with no right contact in the last ten
and no table contact since success. A two-second window can miss a slow but complete handover. Of 183 successes of
the absolute-error head that failed strict success, 120 still held the cube ten seconds later. \emph{Held at $T$}
therefore continues the frozen policy for $T$ seconds after the first success and requires a hold through the final
25 steps ($T=10$\,s for Human, 5\,s for Scripted; Held at 5 and 10\,s agree in 99.5\,\% of episodes). These labels
come from simulator contacts and poses and are used only for evaluation.

\subsection{Baseline configurations}
\label{app:baselines}
All baselines (Fig.~\ref{fig:cost}) were trained with LeRobot 0.6.2 on exactly the memory's 40 Human demonstrations (actions, states and
images match frame by frame), with training seed 0 and the final checkpoint, without checkpoint or hyperparameter
selection. ACT used LeRobot's recipe for this task (ResNet-18, chunk 100, 100\,000 steps, batch 8, learning rate
$10^{-5}$) and Diffusion Policy LeRobot's defaults (ResNet-18, two observation frames, horizon 64, 100 DDPM steps,
100\,000 steps, batch 32, learning rate $10^{-4}$). SmolVLA was fine-tuned from \texttt{smolvla\_base} with its vision
encoder frozen (20\,000 steps, batch 64) and X-VLA from \texttt{xvla-base} with every weight trained (20\,000 steps,
batch 8). The released ACT is LeRobot's \texttt{act\_aloha\_sim\_transfer\_cube\_human}, trained on all 50
demonstrations. $\pi_0$ and $\pi_{0.5}$ fit our GPU only with their VLM frozen, a deviation from openpi's recipe, and
reached 33 and 15\,\% on 200 resets, so we do not report them. Every arm re-plans every 25 steps, as \ours{} does. Run
with their native execution lengths instead (ACT 100, Diffusion Policy 32, X-VLA 30, SmolVLA 50 steps), the arms
changed by $-7.5$, $-3.5$, $+3.0$ and $-5.5$ points, respectively, on 200 resets, so the shared schedule did not hold them back. Policy
calls are timed on a dedicated GPU (batch 1, frame and joints to chunk, 1\,056 calls per process on 176 stored
ALOHA inputs). \ours{}'s call computes the key and head tokens in one encoder pass without diagnostics (p95 2.6\,ms;
4.1\,ms before those cleanups) and replays a captured CUDA graph (1.1\,ms), with bitwise identical actions. The baselines run as LeRobot implements them;
graph replay could shorten them too. In
the closed loop, \ours{} re-plans in 1.3\,ms. The fused three-head ensemble takes 2.78 and 1.54\,ms without and with
graph replay at a GPU peak of 193 and 186\,MB (one head, same session: 2.53 and 1.07\,ms, 139 and 132\,MB; ACT, another
session: 3.53\,ms, 266\,MB) and matches the per-head average to $1.1\cdot10^{-6}$\,rad; Threading's heads with their own
vision branch are averaged one by one (5.8 against 2.1\,ms). Per encoder (24\,s, validation / learned $\beta$, Table~\ref{tab:aloha}):
E1 70.9 / 75.9, E2 62.2 / 64.6, E3 68.1 / 82.3, E4 69.1 / 71.8, E5 83.4 / 81.2 and E6 82.2 / 82.7\,\%; over encoders
the s.d.\ is 8.0 and 6.8 points, over heads 3.8 and 3.4 (variance components). Our Diffusion Policy also behaves partly like a lookup table: on the ten held-out demonstrations, its chunks lie
closer to a stored training chunk than the true motion does in 94\,\% of 160 queries (median 1.33$^\circ$ against
1.80$^\circ$; \ours{} 1.70$^\circ$, 54\,\%), consistent with He et al., though a smoothing regressor could show the
same.

\subsection{Recall bandwidth (\texorpdfstring{$\beta$}{beta}) on Human and Scripted ALOHA}
\label{app:beta}
\begin{table}[ht]
  \centering
  \caption{Effect of $\beta$ on Human and Scripted ALOHA (200 dev.\ resets; three heads per corrected point). Median
  rows averaged ($n_{\mathrm{eff}}$) and 24\,s success alone and with a head; ``halved'' is half of ``valid.'',
  confirmed on 600 fresh resets.}
  \label{tab:beta}
  \footnotesize
  \setlength{\tabcolsep}{2.6pt}
  \begin{tabular}{@{}lrrrrrrrr@{}}
    \toprule
    & \multicolumn{4}{c}{Human (40 teleoperated)} & \multicolumn{4}{c}{Scripted (40 scripted)} \\
    \cmidrule(lr){2-5}\cmidrule(l){6-9}
    readout & $\beta$ & $n_{\mathrm{eff}}$ & alone & +head & $\beta$ & $n_{\mathrm{eff}}$ & alone & +head \\
    \midrule
    top-1                   & $\infty$ & 1     & 61.0 & 38.2 & $\infty$ & 1   & 80.0 & 93.7 \\
    $\beta\times4$          & 80  & 25    & 62.5 & 56.2 & 320 & 6   & 55.0 & 93.7 \\
    $\beta\times2$          & 40  & 92    & 61.5 & 70.2 & 160 & 11  & 56.0 & 93.7 \\
    valid.\ $\beta$         & 20  & 479   & 44.0 & 78.8 & 80  & 23  & 58.5 & 94.7 \\
    halved, $\beta\times\frac12$ & 10 & 2\,454 & 14.0 & \textbf{82.5} & 40 & 53 & 58.0 & \textbf{97.2} \\
    $\beta\times\frac14$    & 5   & 8\,213 & 0.0 & 66.7 & 20  & 143 & 45.5 & 94.8 \\
    \bottomrule
  \end{tabular}
\end{table}

\begin{figure}[ht]
  \centering
  \includegraphics[width=0.9\columnwidth]{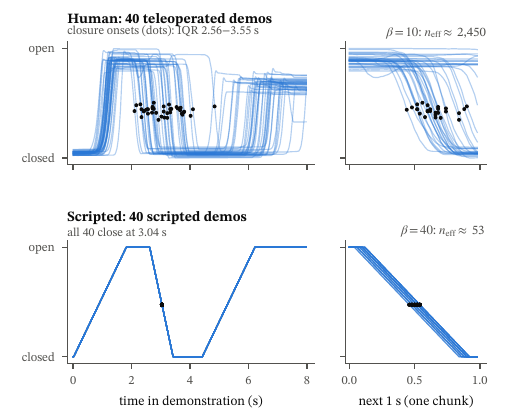}
  \caption{Right-gripper commands of the 40 demonstrations stored in each memory, which show why Human and Scripted
  need different $\beta$. Left: the whole episode, with dots at the closure onset. Right: the next second, which is
  one stored chunk, from the first frame at which the right fingertip is below 6\,cm, i.e., the same situation in
  every demonstration. Scripted futures from that situation are identical up to a few steps, so a few dozen rows
  already form the prototype. Human futures from the same situation disagree on when to close (standard deviation
  0.53\,s), so a smooth prototype takes thousands.}
  \label{fig:demos}
\end{figure}

The validation rule (a grid from 10 to 1\,000, lowest chunk
error of recall alone) picks $\beta=20$ on Human and 80 on Scripted (Table~\ref{tab:beta}). $\beta$ values are not
comparable across encoders, but $n_{\mathrm{eff}}$ is: at the halved settings, Human recall averages about 2\,450
rows, 15\,\% of the bank, and Scripted recall 53, 0.3\,\%.

With a head, Human success peaks near 2\,500 rows (82.5\,\%); relative to the validation choice, making $\beta$ two or
four times sharper costs 8.7 and 22.7 points, while making it four times softer costs 12.2 points (each Holm-separated).
On Scripted, the head corrects every readout from top-1 to 143 rows to 93.7--97.2\,\%. Fig.~\ref{fig:demos} shows why.
The scripted policy executes a single motion: all 40 demonstrations close
the right gripper at the same step (3.04\,s), and from the same state (fingertip first below 6\,cm), all 40 next-second
chunks close within 0.08\,s of one another. Neighboring rows therefore store nearly identical futures, a few dozen rows
already form a clean prototype, and averaging more rows does not further smooth recall's error (Appendix~\ref{app:why}).
Human operators, in contrast, vary in path and timing: their closure onsets span 2.56--3.55\,s (interquartile range), and
from the same state, 26 of the 40 demonstrations close within the next second with a standard deviation of 0.53\,s. A few
rows then reproduce one demonstration's timing, whereas thousands average across operators and halve how rapidly recall's
error changes from frame to frame. Uncorrected recall goes the other way and performs best when sharp on both
datasets (Human 62.5\,\% at 25 rows; Scripted 80.0\,\% at top-1).

A declared test on 600 fresh resets confirmed halving the validation $\beta$ ($+5.1$\pp{} Human, $+1.6$\pp{} Scripted,
Holm $p\le10^{-4}$); neither $\times0.35$ nor $\times0.7$ improved on it. LIBERO-Goal and robomimic were flat across
$\times0.25$--$\times4$, whereas MimicGen gained 8.5--13.2 points at $\beta=1$ (Appendix~\ref{app:ablations}). In
practice, we recommend learning $\beta$ offline from the validation choice on sparse human demonstrations
(Section~\ref{sec:execution}, $\sim$100\,s on ALOHA), and trying a softer $\beta$ in closed loop only where recall does
not carry the policy (retraining only the head, $\sim$70\,s).

\subsection{Why the base matters on ALOHA Human}
\label{app:why}
\begin{figure*}[t]
  \centering
  \includegraphics[width=\textwidth]{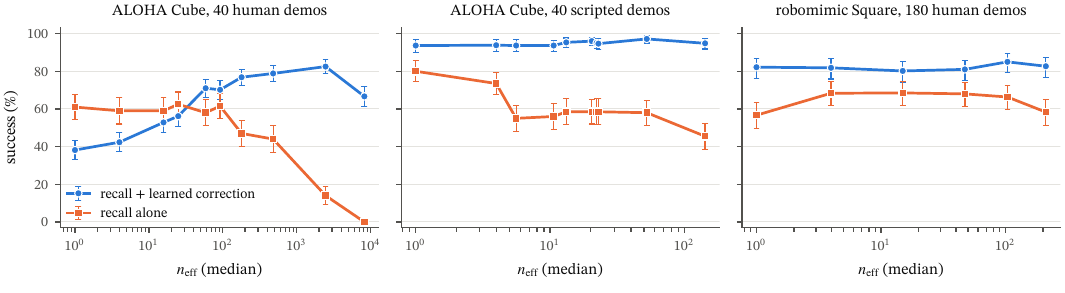}
  \caption{Readout sweeps on three benchmarks: success against the median number of rows recall averages, each
  readout with its own learned correction (blue) and without one (orange). ALOHA: top-1 to top-256 and the whole
  bank at $\beta\times\{4,2,1,0.5,0.25\}$, 200 resets, three heads per corrected point, 95\,\% bootstrap intervals.
  robomimic Square: top-1 to all rows, 200 fresh scenes, three seeds, Wilson intervals. At top-1 the sweep's head
  reads one candidate; Fig.~\ref{fig:mechanism}a controls this. The ALOHA resets were used during
  development, so the sweep is descriptive; the chosen $\beta$ was confirmed on 600 fresh resets.}
  \label{fig:sweep}
\end{figure*}
Fig.~\ref{fig:sweep} repeats the sweep on scripted ALOHA and robomimic Square; only Human shows the reversal. The
top-1 readout switches demonstrations between consecutive held-out frames on 11.8\,\% of frames on Human, 22.9\,\% on
Square and 36.4\,\% on Scripted, so switching alone does not explain it; averaging's effect on the error does. The change of the error per frame,
relative to its size, falls from 0.25 (top-1) to 0.12 (soft) on Human, from 0.66 to 0.57 on Square, and not at all on
Scripted (0.49 to 0.55). Averaging thus makes the base's error systematic on human data. Across fourteen memories,
however, this smoothing separates human from consistent data but does not order the gains (Appendix~\ref{app:rule}).

\subsection{The rule across memories}
\label{app:rule}
\begin{table}[ht]
\caption{Soft minus top-1 base under the head on every memory tested (fresh resets), with the median distance from a
demonstration's grasp to the nearest other one and each base's success under its own heads (first test if two).
ALOHA Human: Table~\ref{tab:replication}. Square, Can: robomimic. Not ablated on LIBERO or the SO-101.}
\label{tab:rule}
\centering\footnotesize
\setlength{\tabcolsep}{2.2pt}
\begin{tabular}{@{}lllrr@{}}
\toprule
Memory & source & spacing & soft / top-1 & soft $-$ top-1 \\
\midrule
ALOHA Human (40) & human & 1.39\,cm & 69.4 / 47.1 & $+22.4$ \\
Square MH, better (90) & human, 2 op. & 0.81\,cm & 63.2 / 55.3 & $+7.9$ \\
Square MH, mixed (90) & human, 6 op. & 0.99\,cm & 32.2 / 25.5 & $+6.7$ \\
Square MH-40 & human, 6 op. & 1.29\,cm & 14.5 / 8.4 & $+6.1$ \\
Square PH-40 & human, 1 op. & 1.52\,cm & 23.9 / 20.5 & $+3.4$ \\
ALOHA Scripted (40) & scripted & 1.49\,cm & 97.2 / 94.5 & $+2.7$ \\
Square PH (180) & human, 1 op. & 0.57\,cm & 84.9 / 82.4 & $+2.5$, $+2.9$ \\
Can MH, mixed (90) & human, 6 op. & 1.54\,cm & 91.5 / 89.1 & $+2.4$ \\
Can MH, better (90) & human, 2 op. & 1.60\,cm & 99.5 / 97.5 & $+1.9$ \\
Can PH (180) & human, 1 op. & 1.09\,cm & 99.0 / 98.4 & $+0.6$ \\
Square MH (270) & human, 6 op. & 0.50\,cm & 51.6 / 51.0 & $+0.6$, $+0.7$ \\
Can MH (270) & human, 6 op. & 0.88\,cm & 95.2 / 96.6 & $-1.4$ \\
MimicGen Square-40 & generated & 0.97\,cm & 20.5 / 22.6 & $-2.1$ \\
MimicGen Square (200) & generated & 0.50\,cm & 50.3 / 53.5 & $-3.2$ \\
\bottomrule
\end{tabular}
\end{table}
The declared prediction (Table~\ref{tab:rule}) was ordinal: within Square, the multi-operator 40-demonstration memory should gain more than
the dense multi-operator, sparse single-operator and sparse generated memories, and all three differences were
separated. Grasp spacing alone does not order the gains: two
90-demonstration Square memories gain 7.9 and 6.7 points at 0.81 and 0.99\,cm. Across all fourteen, gains above 5 points come only from
human data in memories of 40--90 demonstrations; the 270-demonstration multi-operator memories gain nothing, and on Can
every 90-demonstration arm succeeds in at least 89\,\% of scenes. Sparse but consistent data (scripted, or one operator) gain
2.7--3.4 points, and generated data lose a little. On these fourteen memories (declared, offline), the number of
demonstrations orders the gain (Spearman $-0.61$, $p=0.011$), but the base error's roughness ratio (top-1 over soft,
Appendix~\ref{app:why}) does not ($+0.20$; 1.9 on Human, 0.89 on Scripted, 1.15--1.36 on every robomimic human
memory). The head's offline error reduction favors top-1 on all fourteen, and LAC@8~\cite{grachevKNNVLA2026} did not
track the gain either ($+0.49$ over six datasets). Disagreement paired with few demonstrations predicts it
(Appendix~\ref{app:toy}).

\subsection{A controlled toy}
\label{app:toy}
\begin{figure}[t]
  \centering
  \includegraphics[width=\columnwidth]{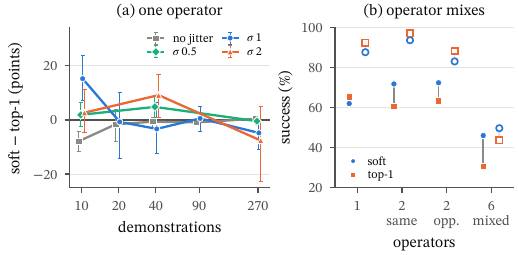}
  \caption{Controlled toy, each base with its own head, three runs $\times$ 500 scenes per point. (a) Soft $-$ top-1
  by jitter $\sigma$ (95\,\% CIs). (b) Success with 40 (filled) and 270 (hollow) demonstrations, $\sigma=1$.}
  \label{fig:toy}
\end{figure}
Disagreement, density, and operator skill covary across our benchmarks, so we separated them in a 2-D
pick-and-place task: pass a wall, grasp an object by closing near it, and carry it to a goal within 15\,s. Scripted
operators follow way-points; a jitter parameter $\sigma$ scales each demonstration's speed, way-point offsets, and pause
before closing; slower, pausing operators stand in for robomimic's weaker ones; and an operator's side of the wall sets
its path. \ours{} reads the z-scored state with the package's head and recipe (22 declared conditions, three runs each,
500 fixed scenes). Without jitter, the soft base never helps ($-7.9$ to $+0.4$ points); with jitter, it helps when
demonstrations are few (post-hoc mean $+6.5$ with at most 40, up to $+15.5$) and does not help when they are plentiful
($-2.1$ with 90 or more; Fig.~\ref{fig:toy}). Six operators of mixed skill preserve the gain at 40 ($+15.5$ [5.6, 24.5])
but lower both bases by 38--49 points at 270 relative to a single operator. Opposite paths do not hurt ($+9.0$ against
$+11.3$) because the state separates them. Four of seven declared predictions held, while the density contrast at
$\sigma=1$ and the offline measures failed.

\subsection{Further ablations}
\label{app:ablations}
Unless stated otherwise, all results below use ALOHA Human with 200 new resets and three heads.
\begin{itemize}
  \item \emph{Key objective} (early protocol: 8\,s success, validation $\beta$, no head, 50 resets $\times$ 3 seeds): SimCLR alone gives 35\,\% with
  top-1 and 0\,\% with soft recall, whereas the action-aware key reaches 51 and 70\,\%.
  \item \emph{Language} (LIBERO-Goal, soft recall, mean $\pm$ s.d.\ over three seeds): $8\pm1$\,\% without
  language, $91\pm3$\,\% fused into the key, and $94\pm2$\,\% with the binding term.
  \item \emph{Candidates.} $K_c=2$ finishes 6.5 fewer handovers [2.3, 10.7] at the same 24\,s success, and $K_c=1$
  and 8 were not better in 40-reset pilots. Without the candidate branch ($K_c=0$), strict success drops by 4.2
  [0.2, 8.3] on ALOHA, while Threading is non-inferior ($+4.7$). Replacing each candidate's chunk with the base chunk
  is not separated ($-2.3$ [$-5.3$, 0.7]).
  \item \emph{Bandwidth on MimicGen.} $\beta=1$ beats the validation choice ($\beta=20$--40) by 8.5--13.2 points.
  Because the same head without query-dependent memory input outperforms the validation-$\beta$ model on Threading and
  Square, most of this gain comes from avoiding the harm of sharp recall.
  \item \emph{Cameras.} Adding the wrist camera adds 20.0 points on robomimic Square and 4.8 on LIBERO-Goal
  (Holm $p\le0.002$).
  \item \emph{Encoder.} Removing crop augmentation costs the head 28.2 points [23.5, 33.0], while leaving soft recall
  alone unchanged. Replacing average pooling with spatial-softmax pooling loses 50.2 points. A DINOv2 ViT-S/14 backbone
  under the same recipe loses 20.5 points on MimicGen Coffee and is not separated on Square or Threading, and 224\,px
  frames produce no change.
  \item \emph{History.} Including past keys adds 22 points to Scripted recall alone, but changes corrected success by
  $-0.3$ and $-2.7$ points.
  \item \emph{Correction inputs on Threading.} Providing the state difference to the readout adds 14.3--17.5 points on
  Threading and has no separated effect on ALOHA, Scripted, or robomimic Square. Training a visual branch for the head
  adds 19.7 points [14.3, 25.0] on Threading over a frozen branch, but costs 9.7--14.7 points on ALOHA, so it is used on
  Threading only.
\end{itemize}

\subsection{Training on thinned memories}
\label{app:thin}
On its own drifting states, the policy's nearest stored row is farther away than in training, so we also trained the
head on memories thinned beyond leave-one-out: in each of eight views per frame, every other demonstration was also
removed with probability 0.25, 0.5 or one drawn from $[0, 0.6]$ (the trainer reproduced \ours{}'s head bit for bit
with nothing extra removed). On 200 fresh resets per dataset and three heads per arm, no rate changed Held at 5\,s, either on Human ($+0.3$, $-1.5$ and $-0.8$ points against 79.8\,\%, Holm
$p=1$) or on Scripted ($-0.8$ to $+0.7$ against 91.3\,\%). On Scripted, every rate missed more first grasps ($+4.0$ to
$+4.8$ points) and lowered 24\,s success by 4.2--5.2 points. Thinning shifted the top-1 similarity the head trains on
only slightly (Human median 0.95 to 0.93--0.94, tenth percentile 0.88 to 0.84--0.87). In the closed loop, by contrast,
the tenth percentile is 0.65, so the far states that matter stay out of reach of this augmentation. A token with the raw top-1 similarity did
not help on Human ($-2.3$ Held) and recovered half the lost grasps on Scripted ($+3.3$).

\subsection{More data}
\label{app:data}
\begin{figure}[ht]
  \centering
  \includegraphics[width=0.9\columnwidth]{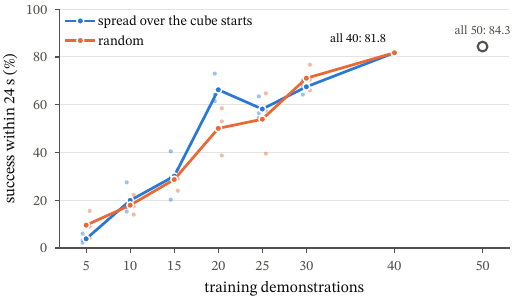}
  \caption{Success on ALOHA cube transfer against the number of human training demonstrations, chosen to spread over
  the cube's start positions (farthest-point order) or at random. Mean of three training seeds (dots), 400 fresh
  resets, the same number of training steps at every size. Hollow: all 50 demonstrations, retrained, on 600 other
  fresh resets.}
  \label{fig:datacurve}
\end{figure}
On robomimic, the unchanged recipe reaches 99.0\,\% on Can and 83.7\,\% on Square
(two cameras, three seeds; the head adds 6.2 and 23.9 points to soft recall), close to BC-RNN's published 98.0 and
82.0\,\%~\cite{mandlekarRobomimic2021} and below Diffusion Policy's on Square~\cite{chiVisuomotorPolicyLearning}
(other protocols; not tested).

To test scaling on generated data, we trained \ours{} on the first 100, 200, 500, and 950 MimicGen Square
demonstrations, reaching 39.7, 50.2, 57.1, and 62.3\,\% on the same 300 resets. Every step is separated (Holm
$p\le0.023$), at about seven points per doubling with no plateau. The head adds 9 points at 100 demonstrations and
is not separated at larger sizes, so the later gains come from the memory. With 200 \emph{human} demonstrations,
\ours{} reaches 82.3\,\% on the same resets, so data source matters more than dataset size.
Doubling the data to 1\,900 with 950 demonstrations that we generated
with MimicGen's released settings did not help ($-1.7$\pp{} [$-4.7$, 1.4], 600 fresh resets). The new
demonstrations cover the same region and are slightly weaker: alone they score 3.7 points less ($p=0.02$).

We observe a sharper threshold when scaling human demonstrations on ALOHA (Fig.~\ref{fig:datacurve}), training \ours{}
on 5 to 40 demonstrations while keeping the number of training steps fixed at each size across three seeds. With 15 or
fewer demonstrations, the policy mostly fails (at most 30\,\%). Success jumps between 15 and 20 demonstrations and
reaches 81.8\,\% with all 40, while retraining on all 50 adds 1.8 points (not separated). Demonstration count matters more
than subset selection: spreading demonstrations over the cube's start positions outperforms random selection only at 20
demonstrations ($+16.2$\pp{}) and hurts at 5 because it prioritizes extreme starts first. Appending demonstrations
40--49 to a frozen encoder and head did not help ($-3.0$\pp{} [$-6.0$, 0.0] at 24\,s), even though 14\,\% of re-plans
retrieved an appended demonstration first.

\subsection{Attempts that did not help}
\label{app:negative}
\begin{table}[ht]
\caption{Changes that were tested in closed loop and not adopted (differences in points against \ours{} on ALOHA
Human unless stated; pilots are marked).}
\label{tab:negative}
\centering\footnotesize
\setlength{\tabcolsep}{3pt}
\begin{tabular}{@{}L{3.2cm}L{4.9cm}@{}}
\toprule
Change & Result \\
\midrule
Flow matching on the memory & no consistent gain over soft recall; worse on Human when started from memory \\
Dynamic readouts (adaptive $\beta$, motion reranking, search, potential fields) & none beat the static readout with the same candidates; 0.6--4.2\,s per re-plan for field rollouts \\
Self-collected success and recovery memories & no gain; simulator-labeled \\
Disagreement gate & $-1.7$, $-3.3$, $+0.3$ on three tasks \\
Directional hold-out & $-8.3$ [$-12.5$, $-4.2$] \\
Local-linear readout & $-16.5$ strict success \\
Bilinear-transduction head; one head per joint (pilots) & $-43.0$; $-33.0$ \\
Soft cross-row consensus base (pilot) & $-8.0$ to $-30.2$ \\
Keypoint branch for the head & edge $+0.0$; overall $-5.0$ [$-8.3$, $-1.7$] Held \\
Training on perturbed queries & Threading $-7.8$; ALOHA $-5.7$ strict success \\
Query-only fine-tuning of retrieval (pilot) & Threading $-17.5$ \\
Unpooled map tokens & ALOHA $-1.7$; Threading $-7.7$ \\
Third attention group & $\pm0$ on Threading, 28\,\% more training \\
Retimed memory (oracle) & $+1.2$ [$-2.5$, 4.8] \\
\bottomrule
\end{tabular}
\end{table}
Beyond Table~\ref{tab:negative}, a head supervised with privileged geometry (cube poses and the arm's Jacobian) gave the
largest gain on ALOHA Human (92.5\,\%) and alone followed a cube outside the demonstrated square, but it needs simulator
labels and fails on scripted demonstrations (32.5\,\%), so it is only an upper reference. Mean key cosine, which had
correlated with encoder quality post hoc, reversed sign on nine crossed encoders (Spearman $+0.47$). A key-uniformity
term hurt Square PH on all six encoders ($-3.7$ [$-5.3$, $-2.1$]), left Threading unchanged, and on ALOHA was separated over encoders only at validation $\beta$ ($+4.9$ [1.0, 9.1]; learned $\beta$: $+2.6$ [$-3.2$, 8.0]). One crop
for both encoder views did not reduce missed grasps ($-0.6$; seed s.d.\ 3.9 to 9.6), and with a fixed epoch 12 it raised
24\,s success by 7.8 [1.4, 13.9] but failed its seed-consistency criterion. Five-step execution with temporal ensembling
added 2.0 points at 24\,s on Table~\ref{tab:aloha}'s heads and 7.0 in strict success on fresh resets, but not on 18
further heads (strict $+1.7$ [$-1.9$, 5.4]), so $E=25$ stays. More conservative single heads lost 4.2--12.1 strict
points, partly because early snapshots of an EMA without warm-up shrink the correction; the EMA after epochs 20--60 lost
6.3--7.8. On Threading, \ours{} with correction vision actively corrects a 4\,mm vertical or lateral needle shift made
about 8\,cm before the ring (signed remainder 8.5 and 2.2\,\%), and its failures are already 1.5 times further off the
ring axis at 8\,cm. Spiking and binary versions of the memory~\cite{wangHASM} preserved the recall (a fully spiking
\ours{} was non-inferior on Scripted), but their spiking head failed on Human; neuromorphic deployment is future work.

\subsection{Retrieval baselines}
\label{app:retrieval}
\begin{table}[ht]
\caption{Retrieval mechanisms on one harness and tuning budget, ALOHA Human, 200 paired resets after 20 tuning
resets per setting; means over three trained seeds where applicable. MCNN and the top-1 residual use \ours{}'s key,
and the others keep the representation of their reference implementation. $^{\dagger}$On \ours{}'s key, 400 other
resets (\ours{}: 85.4\,\%).}
\label{tab:retrieval}
\centering\footnotesize
\begin{tabular}{@{}lrr@{}}
\toprule
Mechanism (selected setting) & 24\,s & strict \\
\midrule
Retrieve-and-play (1-NN) & 39.5 & 33.0 \\
VINN ($k=1$) & 39.2 & 26.2 \\
ALT (our adaptation: image key + robot state) & 39.5 & 27.7 \\
ALT, authors' recipe (images only) & 24.2 & 8.7 \\
BPC mechanism (every step) & 0.0 & 0.0 \\
REGENT mechanism ($\lambda=100$) & 2.8 & 0.8 \\
DARP mechanism ($k=64$, SmolVLM tokens) & 34.8 & 19.5 \\
DARP mechanism, \ours{}'s key ($k=512$)$^{\dagger}$ & 61.5 & -- \\
MCNN mechanism ($\lambda=100$) & 50.0 & 28.5 \\
Top-1 + residual (ReCAP-style) & 51.5 & 38.5 \\
\midrule
\ours{} & \textbf{85.0} & \textbf{76.5} \\
\bottomrule
\end{tabular}
\end{table}
In Table~\ref{tab:retrieval}, MCNN selects 1\,595 of the 16\,000 rows with Neural Gas, blends the nearest memory's action with a bounded network by
$e^{-\lambda d}$ and trains for one million steps. With the selected $\lambda=100$ and a typical key distance of
0.11, the memory's weight is $e^{-11}$, so the tuned MCNN reduces almost to its network. Tuning helped only VINN, by
choosing $k=1$. Retrieve-and-play, REGENT and BPC use frozen ImageNet features with the joint state, VINN uses a
SimCLR key, ALT its contrastive key and DARP frozen SmolVLM tokens. ALT's released code and recipe (images only) score 24.2\,\%, and
the table also reports our stronger adaptation. Under the published protocols
of BPC and DARP (MimicGen Square with 950 demonstrations; PushT), our re-implementations reach 58.8 and 57.2\,\% on
Square against their published 73\,\%, and DARP reaches 15.4 against 70 on PushT. \ours{} reaches 62.3\,\% and a
PushT score of 68.7 (not separated from BPC and DARP on Square). Because our re-implementations miss the published
numbers, the main text relies on the comparison with DARP's own code.

\subsection{The real SO-101 experiment}
\label{app:robot}
\begin{figure}[ht]
  \centering
  \includegraphics[width=0.9\columnwidth]{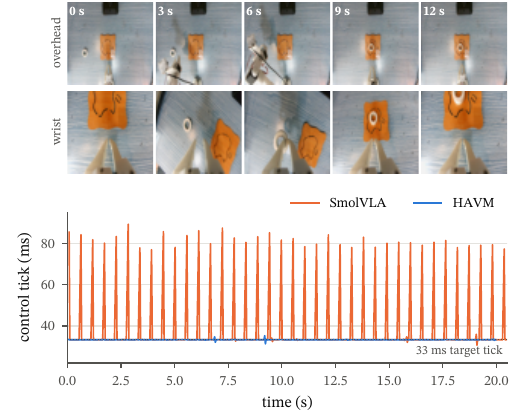}
  \caption{Real SO-101 experiment. Top: an \ours{} trial (overhead and wrist cameras). Bottom: control-tick length in
  a successful trial of each policy at the same spot. SmolVLA's 81\,ms re-plan stalls the 33\,ms loop at every
  re-plan, while \ours{}'s 6\,ms re-plan does not.}
  \label{fig:robot}
\end{figure}
We recorded 50 teleoperated demonstrations of picking tape from five spots with an SO-101 arm
(overhead and wrist cameras, 30\,Hz) and trained \ours{} and SmolVLA~\cite{shukorSmolVLA2025} (fine-tuned from its
pretrained checkpoint, 20\,000 steps) on the same 45. Across two blind rounds at every spot, paired by round and spot,
\ours{} succeeded in 7 of 10 trials and SmolVLA in 8 (exact McNemar $p=1.0$; ten pairs cannot rank the policies or show
equivalence). Both grasp and place at similar times (median first grasp 5.5 and 5.0\,s), but move differently
(Fig.~\ref{fig:robot}): SmolVLA's 81\,ms re-plan (p95) exceeds the 33\,ms tick, stalling the loop at every re-plan with
ten times more direction reversals, whereas \ours{} re-plans in 6\,ms without stalling. Every failure of either policy
occurred at the grasp, with fingers landing on the tape. In \ours{}'s three failures, recall placed 71--81\,\% of its
weight on demonstrations from that spot, retrieving the right trajectories before missing at contact. Across four
sessions and three lighting levels, \ours{} succeeded in 14 of 20 trials.

\subsection{LIBERO details}
\label{app:libero}
In Table~\ref{tab:libero}, SmolVLA, $\pi_0$, OpenVLA-OFT, and \ours{} read third-person and wrist cameras,
proprioception, and the instruction (SmolVLA's \texttt{lerobot/smolvla\_libero} setup: 256\,px cameras and an 8-D
state). In LeRobot's LIBERO harness, Object scenes render darker than their demonstrations (mean pixel 102 vs.\ 137),
so Object is not compared like-for-like. Mixing the other 30 tasks costs at most a few points (LIBERO-Goal: 1.8 [0.2,
3.4] below the 10-task memory). Probes use 20 initial states per task, and closed-loop control adds 13 to 43 points over
open-loop replay.

\subsection{VLA comparison on LIBERO}
\label{app:vla}
The VLA baseline is SmolVLA (0.45\,B parameters), fine-tuned by its authors on the four LIBERO suites (1\,693
demonstrations, matching our memory; checkpoint \texttt{lerobot/smolvla\_libero}). It runs in \ours{}'s harness with its
own pre- and post-processors, paired with \ours{}'s episodes by task and initial state. We required at least 80\,\% on
original instructions; with 50-action chunks it scored 68.75\,\%, so we selected among the two official checkpoints and
re-planning every 1, 10, or 50 steps on development states 40--42 (best: LIBERO checkpoint, every 10 steps, 80.0\,\%;
83.25\,\% on test states). Primary contrasts (Holm over two) compare the drop differences on Goal rewrites and 6\,cm
shifts, paired by episode.

\subsection{OCBench}
\label{app:ocbench}
OCBench~\cite{parkOCBench2026} provides scripted demonstrations with low-dimensional state in any quantity. With 1\,000
demonstrations, its behavior-cloning MLP (two million steps) reaches 72.8/52.6/13.0\,\% on three tasks versus
43.4/7.0/2.6\,\% for state-keyed \ours{}, which grasps with a misaligned wrist twice as often at similar offline error.
With images and 100 demonstrations, both stay near the floor; with abundant scripted data, parametric capacity pays off.

\subsection{Instruction rewrites}
\label{app:rewrites}
Verb substitutions replace verbs in all 40 tasks while preserving object names and spatial relations; Goal sentence
rewrites also alter word order, possessives, prepositions, capitalization, and punctuation (one fixed rewrite per task
and condition, e.g., ``pick up \dots\ and place it in the basket''$\to$``lift \dots\ and put it in the basket''). Both
runs share the model (seed 0), initial states 0--9, and goal check, using exact McNemar tests (Holm over five) with
intervals over task--state pairs. One rewrite per task cannot capture all phrasings. Object shifts move every movable
object (not fixtures) 3 or 6\,cm horizontally in a direction drawn per task and initial state before letting the scene
settle; these perturbations are ours, not LIBERO-PRO's or LIBERO-Plus's.

\end{document}